\PassOptionsToPackage{dvipsnames}{xcolor}

\documentclass[final]{l4dc2025}
\usepackage{times}       
\usepackage{amsmath}     
\usepackage{amssymb}     
\usepackage{multirow}
\usepackage[separate-uncertainty=true,number-unit-product={ }]{siunitx}%
\usepackage{caption}

     \usepackage{xcolor}

    \definecolor{matplotlibpurple}{RGB}{148, 0, 211}
    \definecolor{matplotlibblue}{RGB}{31, 119, 180}
    \definecolor{matplotlibgreen}{RGB}{44, 160, 44}
    \definecolor{matplotliborange}{RGB}{255, 127, 14}
    \definecolor{matplotlibred}{RGB}{214, 39, 40}

\title[Dynamic Safety Shield for Navigation Tasks]{ A Dynamic Safety Shield for Safe and Efficient Reinforcement Learning of Navigation Tasks}

\author{%
 \Name{Murad Dawood} \Email{dawood@cs.uni-bonn.de}\\
 \AND
 \Name{Ahmed Shokry} \Email{shokry@cs.uni-bonn.de}\\
  \AND
 \Name{Maren Bennewitz} \Email{maren@cs.uni-bonn.de}\\
 \addr Humanoid Robots Lab, University of Bonn, Germany%
}

\begin{document}
\maketitle

\begin{abstract} 

Reinforcement learning (RL) has been successfully applied to a variety of robotics applications, where it outperforms classical methods. However, the safety aspect of RL and the transfer to the real world remain an open challenge.
A prominent field for tackling this challenge and ensuring the safety of the agents during training and execution is safe reinforcement learning. 
Safe RL can be achieved through constrained RL and safe exploration approaches.
The former learns the safety constraints over the course of training to achieve a safe behavior by the end of training, at the cost of high number of collisions at earlier stages of the training.
The latter offers robust safety by enforcing the safety constraints as hard constraints, which prevents collisions but hinders the exploration of the RL agent, resulting in lower rewards and poor performance.
To overcome those drawbacks, we propose a novel safety shield, that combines the robustness of the optimization-based controllers with the long prediction capabilities of the RL agents, allowing the RL agent to adaptively tune the parameters of the controller. Our approach is able to improve the exploration of the RL agents for navigation tasks, while minimizing the number of collisions. 
Experiments in simulation show that our approach outperforms state-of-the-art baselines in the reached goals-to-collisions ratio in different challenging environments. The goals-to-collisions ratio metrics emphasizes the importance of minimizing the number of collisions, while learning to accomplish the task.  Our approach achieves a higher number of reached goals compared to the classic safety shields and fewer collisions compared to constrained RL approaches. Finally, we demonstrate the performance of the proposed method in a real-world experiment.
\end{abstract} 

\section{Introduction}
\label{sec:intro}

In the last decade, reinforcement learning (RL) has been shown to outperform classical methods in navigation tasks [\cite{xu2023benchmarking, he2024agile}]. This is mainly due to the fact that RL approaches are able to directly map sensory information into actions, enabling them to learn in complex scenarios that would otherwise necessitate lots of engineering efforts. Nevertheless, the safety aspect of RL and the transfer of the trained policies to the real-world remain challenging. 

Safe RL has been proposed to ensure the safety of the RL agents during training and execution and two main directions have emerged in the field. First, constrained reinforcement learning: this approach formulates the RL as a constrained Markov decision problem to minimize safety constraints violations during training [\cite{achiam2017constrained, tessler2018reward, stooke2020responsive, aghaexploring}].
This method trains an additional network to estimate the cost of constraint violations. The policy is then trained to maximize rewards while minimizing those costs. Although these methods often achieve safe behavior by the end of training, they tend to violate constraints frequently during early learning. 
Additionally, the end-to-end nature of these policies makes their transfer to the real-world even more challenging due to the sim-to-real gap.

The second approach to achieve safe RL is safe exploration. In addition to the RL agent responsible for achieving the task (task agent), a safe policy, also referred to as a safety shield, is introduced to ensure the safety of the task agent during training and execution [\cite{wabersich2021probabilistic, dalal2018safe, zhang2023spatial}]. The safety shield utilizes knowledge about the dynamics of the robot, without having knowledge about the dynamics of the environment, to override unsafe actions from the task agent during training and execution. Although, this approach is more reliable and suffers from less constraints violations, it tends to overly restrict the exploration process, which consequently leads to less rewards achieved by the RL agents. This is caused by the formulation of the safety shields, which focus mainly on collisions avoidance and perform with limited prediction horizons.

\begin{figure}[t!] 
    \centering
    \includegraphics[width=0.65\linewidth]{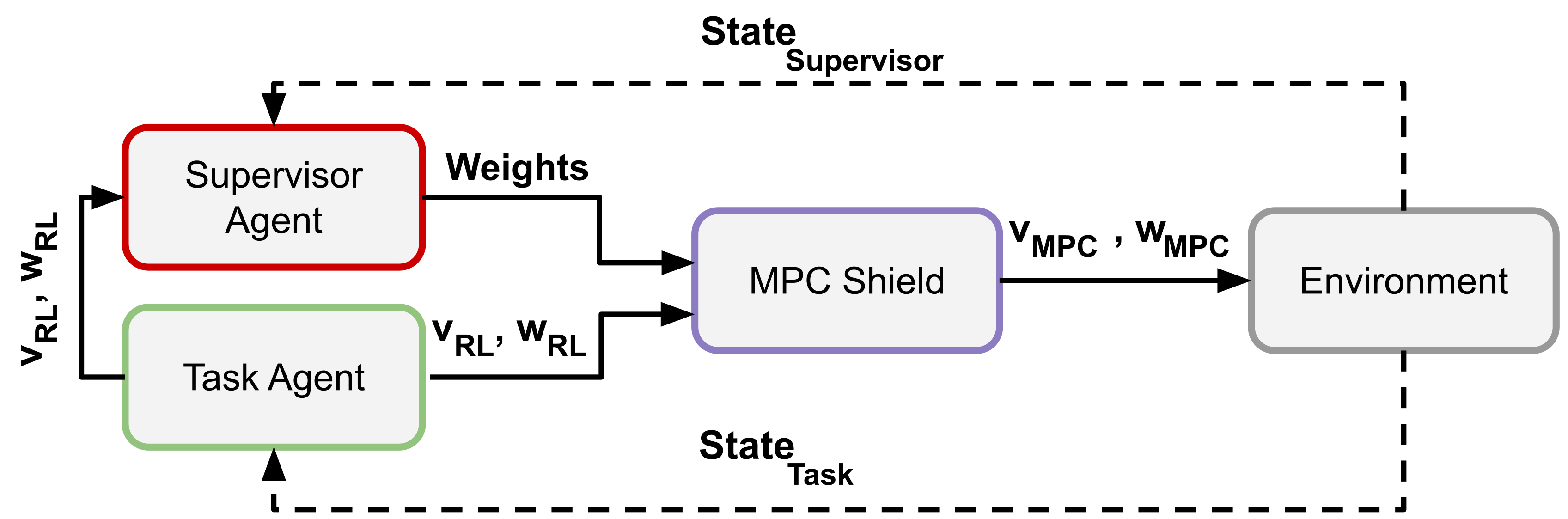}
    \caption{\small Architecture of our approach. The task agent (green) is responsible for learning the navigation task. The agent receives the \textbf{$State_{Task}$} from the environment and outputs the linear and angular velocities~(\textbf{$v_{RL}, w_{RL}$}). The supervisor agent (red) receives the \textbf{$State_{Supervisor}$} from the environment and outputs the \textit{Weights} for aligning the MPC-shield's actions with the task agent's actions, and the weights of the constraints. The MPC shield solves the optimal control problem (Eq.~\ref{eq:mpc_mod}) using the all mentioned weights to find the safe actions \textbf{$v_{MPC}, w_{MPC}$}.
    }
    \label{fig:pipe_coop}
    \vspace{-15pt}
\end{figure}

To address these shortcomings, we propose integrating the long-horizon predictive capabilities of reinforcement learning with the robustness of optimization-based controllers, see Fig. \ref{fig:pipe_coop}. Specifically, we combine a model predictive control (MPC)-based safety shield with an RL agent to provide enhanced guidance to the task agent. This agent is referred to as the supervisor agent and is responsible for dynamically adjusting the weights of the constraints based on the current observations. Additionally, to ensure that the safety shield does not overly constrain the task agent's exploration, the supervisor agent also adjusts the weights to align the shield's actions with those of the task agent. The supervisor agent is responsible for avoiding collisions while matching the actions of the task agent, without having knowledge about the main task of the task agent. A clear advantage of not informing the supervisor agent about the main task is to minimize the exploration needed for the supervisor agent to converge. This results in a small number of constraints violations, while not affecting the exploration of the task agent, as we show in the experiments.

In this work, we focus specifically on safety in navigation tasks, where the task agent controls the robot's linear and angular velocities. Thus, the considered constraints are related to obstacle avoidance and constraints violations are collisions done by the learning robot. 

In summary, the main contributions of our work are: \textbf{(i)} A novel safety shield for navigation tasks that combines the robustness of the MPC shields with the long-horizon capabilities of the RL agents, by allowing an additional RL agent to dynamically adjust the weights of the constraints and the weights for matching the task agent's actions. \textbf{(ii)} A supervisor RL agent without access to goal-related information to dynamically adjust the MPC shield online. This goal-independent training results in small number of collisions as we show in the ablation study. \textbf{(iii)} We show over several simulation environments of increasing difficulties the superiority of our approach over several state-of-the-art baselines w.r.t. the reached goals-to-collisions ratio. Finally, we demonstrate the performance of our approach in real-world scenarios.

\section{Related Work}
\label{sec:related}
\vspace{-8pt}
\subsection{Constrained Reinforcement Learning} Several studies proposed formulating the RL task as a constrained Markov decision problem, to take the constraints into consideration during training. In [\cite{achiam2017constrained}] the authors modified the trust-region policy optimization (TRPO) method to include the constraints on expected costs in the update rule of the algorithm. The study [\cite{ray2019benchmarking}] introduced the Lagrangian TRPO and Lagrangian proximal policy optimization methods, and showed that constrained RL methods achieve fewer constraints violations compared to the unconstrained RL at the cost of reduced rewards. However, a common issue in the Lagrangian constrained RL methods was instability during training. In [\cite{stooke2020responsive}] the authors addressed this instability by introducing PID control over the Lagrangian update and showed improved performance. [\cite{thananjeyan2021recovery}] suggested pre-training a separate recovery policy along with a critic that learns the constraints violations offline then continues the training online. The pre-training phase requires a replay buffer that demonstrates examples of constraints violations. [\cite{sootla2022saute}] proposed using a safety budget, which represents the remaining allowable constrains violation,  instead of the cost to reduce constraints violations. [\cite{aghaexploring}] combined model-based constrained RL with the recovery policy from [\cite{thananjeyan2021recovery}] and showed reduction in the constraint violation compared to several model-free constrained RL baselines. These methods learn the constraints during training to decrease the rate of constraints violation over the course of training, which results in high number of constraints violations during the initial training stages.
\vspace{-10pt}
\subsection{Safe Exploration} On the other spectrum of safe RL lie safe exploration approaches, which promote safer exploration by eliminating actions that would lead to unsafe states. [\cite{wabersich2021predictive}] proposed using MPC-based safety shield to override unsafe actions from the RL agent, where the safe states were formulated as hard constraints. [\cite{dalal2018safe}] pre-trained a neural network to act as the safety shield, and added it to the policy network that is trained from scratch. [\cite{carr2023safe}] used a state estimator in addition to the safety shield to improve the agent safety. [\cite{zhang2019mamps}] utilized predictive safety shields to ensure the safety of the agents in multi-agents scenarios. [\cite{zhang2023spatial}] used control barrier functions [\cite{ames2019control}] as a shielding mechanism to ensure the safety of the agents. 
Different from these methods, we take advantage of the RL long-horizon prediction capabilities to tune the safety shield used in training the RL task agent. This leads to a less constrained shield that does not hinder the exploration of the task agent as we show in the experiments. Moreover, softening the hard constraints and tuning their weights online leads to less failures of the MPC solver, where the MPC is unable to find a solution,  which commonly happens when considering several constraints as in [\cite{brito2021go, dawood2025ral}].
\section{Our Approach}
\label{sec:approach}
In this work, we propose a novel safety shield that is tuned online using an RL agent, i.e., the supervisor agent, to ensure the safety of the task agent, see Fig.\ref{fig:pipe_coop}. The shield along with the supervisor agent do not have access to goal-related information, which is handled by the task agent. Our aim is to develop a safety shield that minimizes the number of collisions, while not hindering the exploration process of the task agent.

First, we introduce the considered navigation task. We then explain the dynamic safety shield and the differences from the classic MPC-safety shield [\cite{wabersich2021predictive, dawood2025ral}]. Afterwards, we introduce the supervisor RL agent responsible for adjusting the shield online. Finally, we introduce the task agent used for solving the task.

\vspace{-5pt}
\subsection{Navigation Task}
We focus on the navigation task, where the RL task agent should learn how to navigate to goals, as fast as possible, without collisions in unknown environments. We assume that the robot is equipped with a lidar sensor and is given the relative distance and heading to the goal. The lidar sensor provides 100 equidistant beams over a range of $360^\circ$, giving obstacle information around the robot. The task agent uses this information to produce the linear and angular velocities for the robot.
\vspace{-5pt}

\subsection{Dynamic Model Predictive Control Safety Shield}
The MPC shield requires a mathematical model of the robot to predict the future states based on the current state and the calculated control sequence. An optimization control problem is solved at each time step to find the control sequence that minimizes a predefined cost function. Only the first control action in the control sequence is applied to the robot. It is important to note that the mathematical model represents the kinematics of the robot only and does not include the dynamics of the world model as in model-based RL [\cite{huang2023safe, aghaexploring}].

\textbf{Prediction Model: }The discrete-time model of the robot is:
$\textbf{x}_{t+1} = \textbf{x}_{t} + \begin{bmatrix} \cos \theta_{t} & 0 \\ \sin \theta_{t} & 0 \\ 0 & 1 \end{bmatrix} \textbf{a}_{t} \Delta t$, where $\textbf{x} = [x, y, \theta]^{T}$ represents the state of the robot, which is the position and heading relative to the starting position. $\textbf{a} = [v, \omega]^{T}$ is the action, which is the linear and angular velocities of the robot.

\textbf{Optimal Control Problem (OCP): }To ensure the safety of the robot while minimizing the intervention of the safety filter, the optimization problem at time step $t$ is formulated as:
\begin{subequations} {
\begin{align}
 \underset{\substack{\textbf{x}_{t:t+T|t},\\ \textbf{a}_{t:t+T-1|t}}}
 {\mathrm{min}}
& \lVert \textbf{a}_{\mathit{RL}} - \textbf{a}_{t|t} \rVert^2_{\mathit{R_{0}}} + \sum_{k=1}^{T-1}  \lVert \textbf{a}_{t+k|t} \rVert^2_{\mathit{R}} \label{eq:cost}\\
 \text{s.t.} \quad 
& \textbf{x}_{t|t} = \textbf{x}_{t},\label{eq:co1}\\
 & \textbf{x}_{t+k+1|t} = f(\textbf{x}_{t+k|t}, \textbf{a}_{t+k|t}), \forall k= 0,1,...,T-1 \label{eq:co2}\\
 &  \textbf{a}_{t+k|t} \in \textbf{A},  k= 0,1,...,T-1 \label{eq:co3}\\
&  \textbf{x}_{t+k|t} \in \textbf{X}, \forall k= 0,1,...,T-1, \label{eq:co4}\\
&   {\mathit{dist}}^{t+k|t}_{\mathit{obst}} > \delta, \forall obst= 1,2...,M, \forall k= 0,1,...,T-1, \label{eq:co5}
\end{align}}\label{eq:mpc}
\end{subequations}
where $T$ represents the prediction horizon. The weighted Euclidean norm, represented by \( \lVert \cdot \rVert^2_R \), is defined as \( \lVert \textbf{x} \rVert^2_R = \textbf{x}^\mathsf{T} R \textbf{x} \), where \( R \) is a positive definite weighting matrix. The notation ${t+k|t}$ indicates predictions at time $t+k$, assuming the current time is $t$. 
The first term in (\ref{eq:cost}) measures the deviation between the task agent's proposed action $\textbf{a}_{\mathit{RL}}$ (Action$_{RL}$) and the initial MPC action $\textbf{a}_{t}$ (Action$_{\mathit{MPC}}$).
The second term minimizes the magnitude of future control signals ${\textbf{a}}_{\mathit{t+k}}$, promoting smoother transitions. Weight matrices $R_{0}$, $R$ are used to optimize the performance of the shield and are manually tuned. 
Constraints (\hyperref[eq:co1]{\ref*{eq:co1}--\ref*{eq:co5}}) are hard constraints:
(\ref{eq:co1}) ensures the initial state of the model matches the actual state of the robot,
(\ref{eq:co2}) enforces the robot's dynamic model,
(\ref{eq:co3}) and (\ref{eq:co4}) bounds the actions and states within the predefined limits, where $\textbf{X}$ and $\textbf{A}$ denote the allowable sets of states and controls, respectively.
(\ref{eq:co4}) enforces the distance between the robot and the nearest $M$ obstacles to be larger than a safety threshold distance $\delta$. $M$ is a design parameter representing the maximum number of obstacles considered by the MPC. Since the MPC process raw lidar data directly, the lidar data is divided into M equal sectors and only the beam with the smallest distance in each sector is considered as obstacle. In this work, we use $M=4$.

The main drawbacks of the above formulation are that considering several obstacles as hard constraints in the MPC problem can often lead the solver to fail to find a feasible solution online as in [\cite{brito2021go}], and manually tuning the weight matrix $R_{0}$ to achieve a reasonable performance is a time-consuming process. That is why we propose modifying the OCP as follows:
\begin{subequations} {
\begin{align}
 \underset{\substack{\textbf{x}_{t:t+T|t},\\ \textbf{a}_{t:t+T-1|t}}}
 {\mathrm{min}}
& \lVert \textbf{a}_{\mathit{RL}} - \textbf{a}_{t|t} \rVert^2_{\mathit{\color{PineGreen}R_{0}}} + \sum_{k=1}^{T-1}  \lVert \textbf{a}_{t+k|t} \rVert^2_{\mathit{R}} \color{PineGreen} + \sum_{obst=1}^M \dfrac{\omega_{obst}}{{\mathit{dist}}^{t+k|t}_{\mathit{obst}}} \label{eq:cost_mod}\\
 \text{s.t.} \quad 
& \textbf{x}_{t|t} = \textbf{x}_{t},\label{eq:co6}\\
 & \textbf{x}_{t+k+1|t} = f(\textbf{x}_{t+k|t}, \textbf{a}_{t+k|t}), \forall k= 0,1,...,T-1 \label{eq:co7}\\
 &  \textbf{a}_{t+k|t} \in \textbf{A},  k= 0,1,...,T-1 \label{eq:co8}\\
&  \textbf{x}_{t+k|t} \in \textbf{X}, \forall k= 0,1,...,T-1, \label{eq:co9}
\end{align}}\label{eq:mpc_mod}
\end{subequations}
where the green terms in the cost function \ref{eq:cost_mod} indicates our modifications. The weight matrix $R_{0}$ is now tuned online, while the obstacle avoidance terms scaled by the weights $\omega_{obst}$ are now added to the cost function. However, setting these weights equally often leads the robot to being stuck when surrounded by obstacles [\cite{dawood2025ral}]. That is why we learn these weights online based on the observations of the supervisor agent which we discuss in the next section. We do not tune the $R$ matrix online, We do not tune the R matrix online, as its purpose is to reduce the control effort over the rest of the prediction horizon.

\vspace{-5pt}

\subsection{Supervisor Reinforcement Learning Agent}

To adjust the MPC shield online, we use a soft-actor-critic~[\cite{haarnoja2018soft}] (SAC) agent, which we call the supervisor agent. The role of the supervisor agent is to modify the weights of the obstacle terms used in the cost function of the MPC shield \ref{eq:cost_mod}, in addition to adjusting the weights for matching the actions from the task agent, see Fig. \ref{fig:pipe_coop}. We adapt the Markov decision process~(MDP), which is described by a tuple $M$: ~($S$, $A$, $R$, $P$, $\gamma$). Where $S$ is the set of states, $A$ is the set of actions, $R(s,a)$ is the reward function, $P(s'|s, a)$ is the transition probability, and $\gamma$ is the discount factor. An agent in state $s \in S$ takes an action $a \in A$ resulting in the next state $s' \in S$, which is rewarded by reward~$r$ and discounted by factor $\gamma$. The action $a$ is chosen according to a policy $\pi$ that determines for each state which action the agent will take. The transition from state $s$ to state $s'$ upon taking action~$a$ is determined by the transition probability $P$.

The \textbf{observation space} for the supervisor agent includes the lidar data, the action from the task agent, the previous action from the MPC shield, and the relative angles and distances of the closest $M$ obstacles fed to the MPC, which empirically improves performance. The \textbf{action space} of the agent includes the two weights for penalizing the deviation between the task agent and MPC shield, in addition to the $M$-dimensional weights corresponding to the obstacles considered by the MPC. We formulate the \textbf{reward function} as follows:
\begin{equation}
\begin{aligned}
& r = \begin{cases}
-r_{collision} &\text{if collision or stuck},\\
- min\_dist_{obst} \cdot \rVert Action_{RL} - Action_{MPC} \rVert^2  &\text{otherwise}
\end{cases}
\end{aligned}
\end{equation}
Since the supervisor agent is not responsible for the task completion, we do not give rewards for reaching the goal.
Instead, we give the agent a large penalty $r_{collision}$ if the robot collides or if the robot is not moving for several consecutive steps (stuck), or a continuous penalty as a function of the distance to the single nearest obstacle and the difference between the action of the task agent and the MPC shield. That is, the further the robot is from the obstacles, the higher the penalty is for not matching the action from the task agent. As the robot approaches an obstacle, the penalty decreases, prioritizing safety over action alignment. This encourages the supervisor agent to follow the task agent's actions when far from obstacles and adjust them near obstacles to prevent collisions. 

\subsection{Task Agent} 
The task agent is also a SAC agent, whose observation space includes the lidar data, the previously taken action by the robot ($v_{MPC}$ and $\omega_{MPC}$), and the relative angle and distance to the goal location. The action space consists of $v_{RL}$ and $\omega_{RL}$. The reward function is as follows:
\begin{equation}
r = \begin{cases} 
r_{goal} & \text{if goal is reached}, \\ 
({goal\_dist}_{t-1} - {goal\_dist}_{t}) & \text{otherwise}.
\end{cases}
\end{equation}
where the agent receives a large reward $r_{goal}$ for reaching the goal and a continuous term that rewards the progress towards the goal. We do not penalize the collisions for this agent and rely on the supervisor instead to eliminate them.

\section{Experiments}
This work focuses on achieving safe reinforcement learning in navigation tasks by minimizing collisions without restricting the task agent's exploration. The experiments are designed to: (1) evaluate the impact of the dynamic shield on RL training, (2) compare the proposed approach with baseline methods in simulations, (3) analyze the effect of excluding task-dependent information for the supervisor, and (4) illustrate how the dynamic shield adjusts the constraints' weights on the real robot.
\begin{figure}[t!]
    \centering
    \subfigure[Environment 1 ]{
        \includegraphics[width=0.25\textwidth]{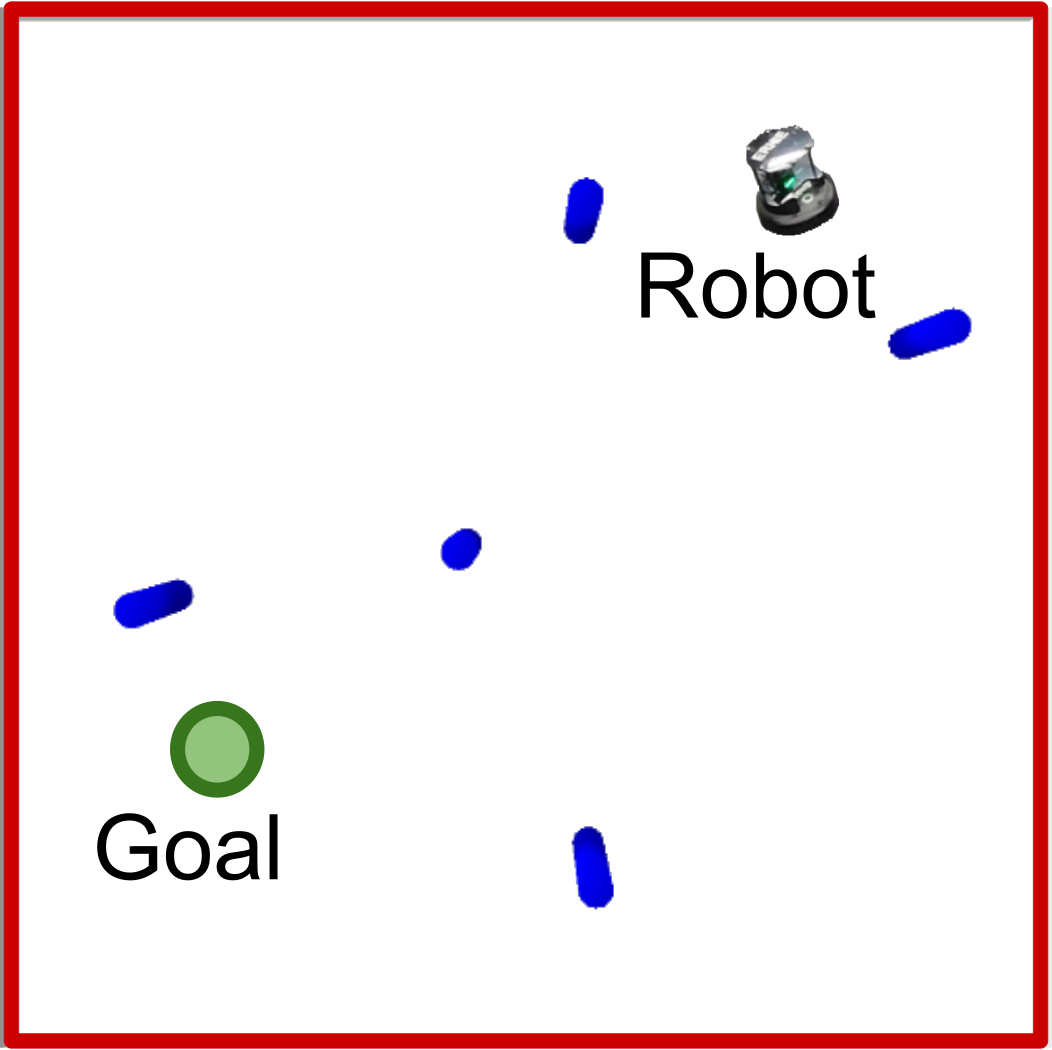}
        \label{fig:env0}
    }
     \subfigure[Environment 2]{
        \includegraphics[width=0.25\textwidth]{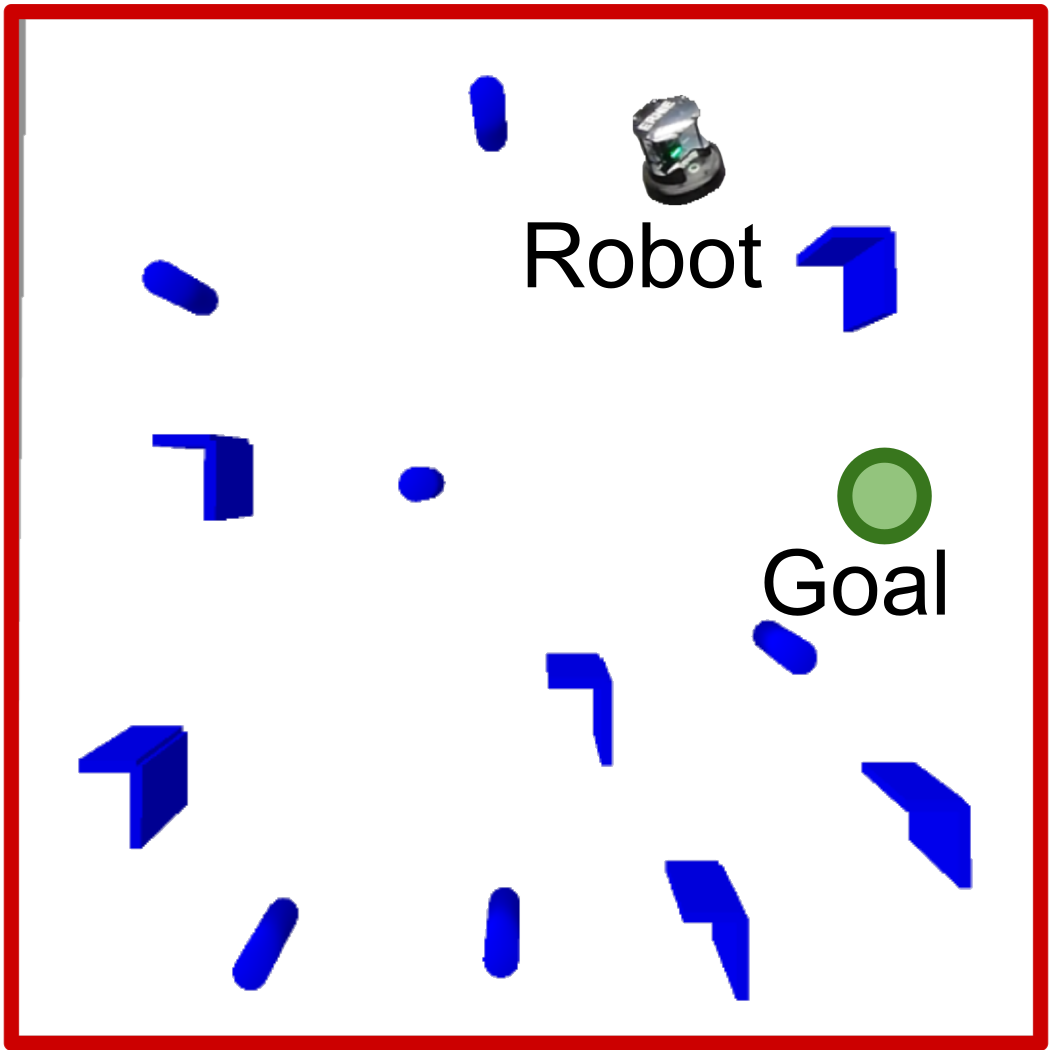}
        \label{fig:env1}
        }
    \subfigure[Environment 3]{
        \includegraphics[width=0.25\textwidth]{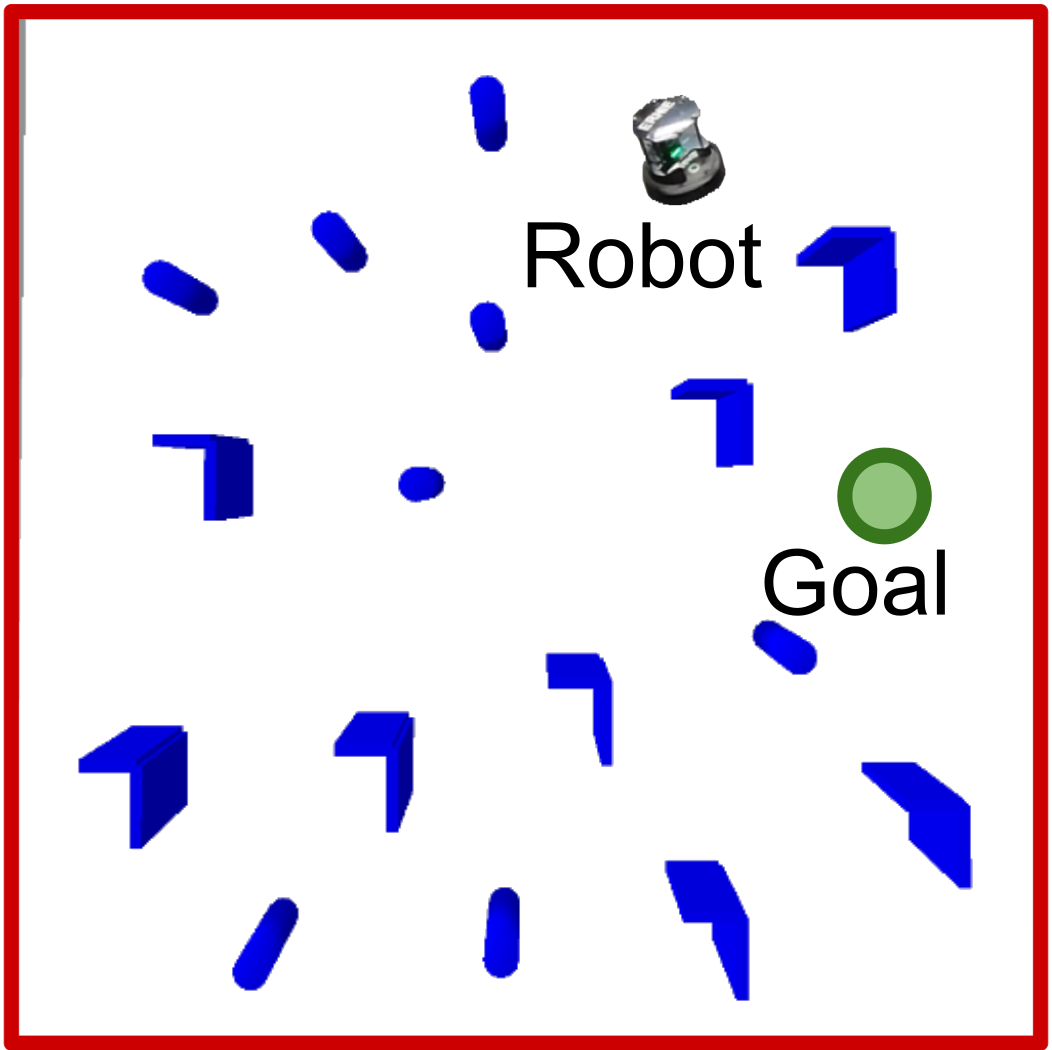}
        \label{fig:env2}
    }

\caption{\small Environments used in the experiments, Fig.\textbf{a} environment with five pillars (blue), and Fig.\textbf{b} environment, which contains six pillars and six L-shaped walls (blue). Fig.\textbf{c} environment with eight pillars and eight L-shaped walls. All the obstacles are placed randomly at the beginning of each episode.}
    \label{fig:envs}
    \vspace{-18pt}
\end{figure}

\vspace{-10pt}
\subsection{Baselines}
For a fair comparison, we chose the baselines such that all of them are model-free RL, all baselines are trained from scratch and do not require pre-training, the baselines include variants of constrained-RL and safe exploration methods: 
\textbf{ (i) SAC~[\cite{haarnoja2018soft}]}: Unconstrained SAC.
\textbf{ (ii) SAC-Lagrangian~[\cite{ray2019benchmarking}]  (SAC\textunderscore LAG):} The Lagrangian variant of the SAC is a constrained-RL approach which uses two additional critic networks ($Q_{cost}$) to estimate the costs and optimizes the Lagrange multiplier online to balance the goal reaching with the rate of collisions. 
\textbf{ (iii) SAC-PID~[\cite{stooke2020responsive}] (SAC\textunderscore PID)\footnote{We used the same code as in [\cite{omnisafe2023}] for SAC\textunderscore LAG and SAC\textunderscore PID}:} A PID controller [\cite{aastrom2006advanced}] is used along with the SAC-Lagrangian to update the Lagrange multiplier. The method has been shown to stabilize the training of the costs' critic ($Q_{cost}$) and has been commonly used as a baseline in safe RL studies. 
\textbf{(iv) MPC Safety Shield~[\cite{dawood2025ral}] (MPC\textunderscore Tuned):} The MPC-based safety shield is a pre-tuned safety shield that ensures zero collisions in navigation scenarios. It has already been used in safe multi-agent RL and is classified as a safe exploration method.
\begin{figure}[t!]
    \centering
        \textcolor{matplotlibpurple}{\rule[0.5ex]{0.5cm}{4pt}} Ours \hspace{1em}
        \textcolor{matplotlibblue}{\rule[0.5ex]{0.5cm}{4pt}} SAC \hspace{1em} 
        \textcolor{matplotlibgreen}{\rule[0.5ex]{0.5cm}{4pt}} SAC\_LAG \hspace{1em}
        \textcolor{matplotliborange}{\rule[0.5ex]{0.5cm}{4pt}} SAC\_PID \hspace{1em}
        \textcolor{matplotlibred}{\rule[0.5ex]{0.5cm}{4pt}} MPC\_TUNED \\

    \makebox[0.315\textwidth]{\textbf{Environment 1}} \hfill
    \makebox[0.315\textwidth]{\textbf{Environment 2}} \hfill
    \makebox[0.315\textwidth]{\textbf{Environment 3}} \\
    \vspace{5pt} 
    \subfigure[]{
        \includegraphics[width=0.315\textwidth, trim={33pt 0pt 55pt 0pt}, clip]{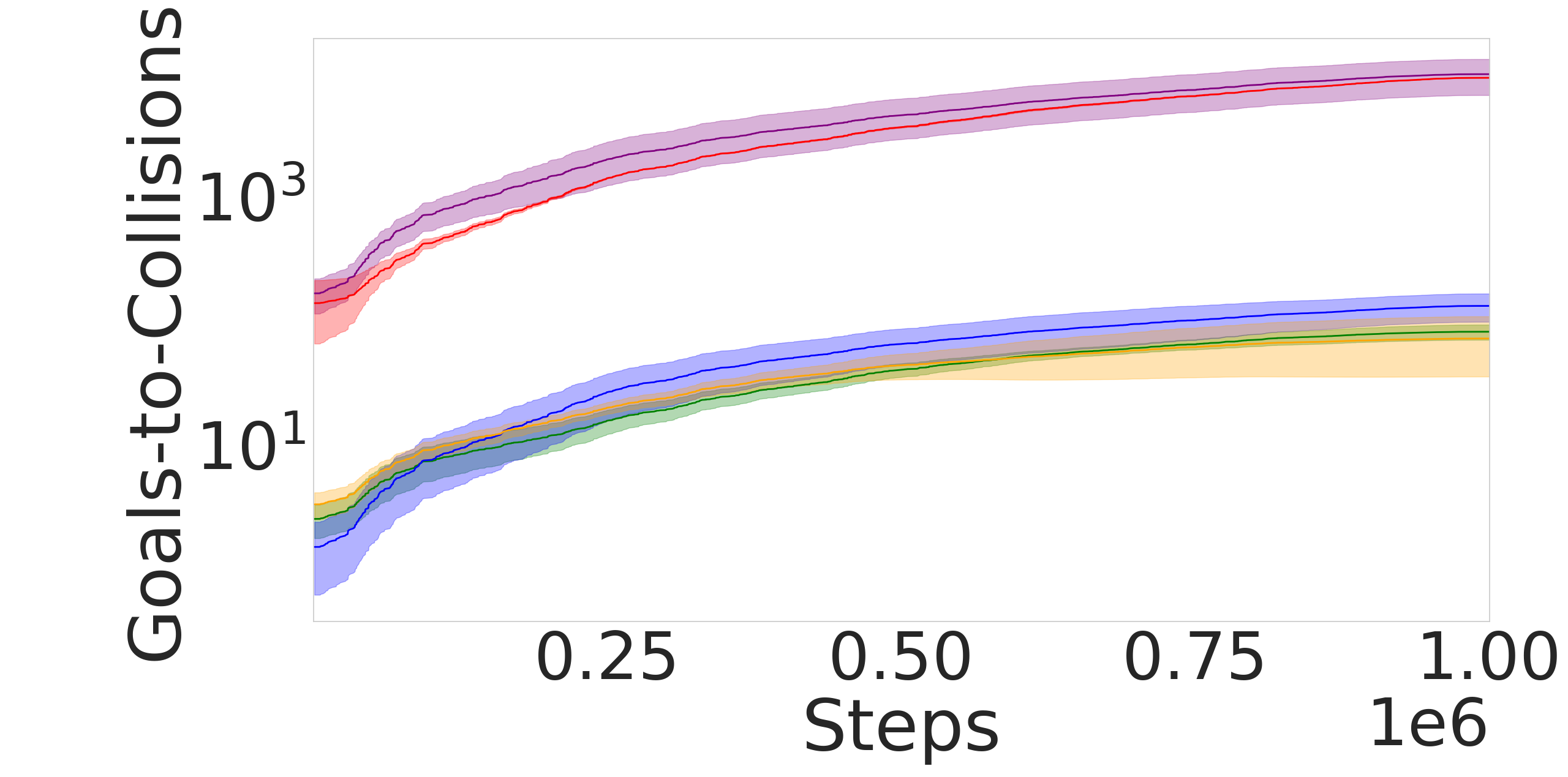}
        \label{fig:ratio_0}
    }
    \hfill
    \subfigure[]{
        \includegraphics[width=0.315\textwidth, trim={35pt 0pt 55pt 0pt}, clip]{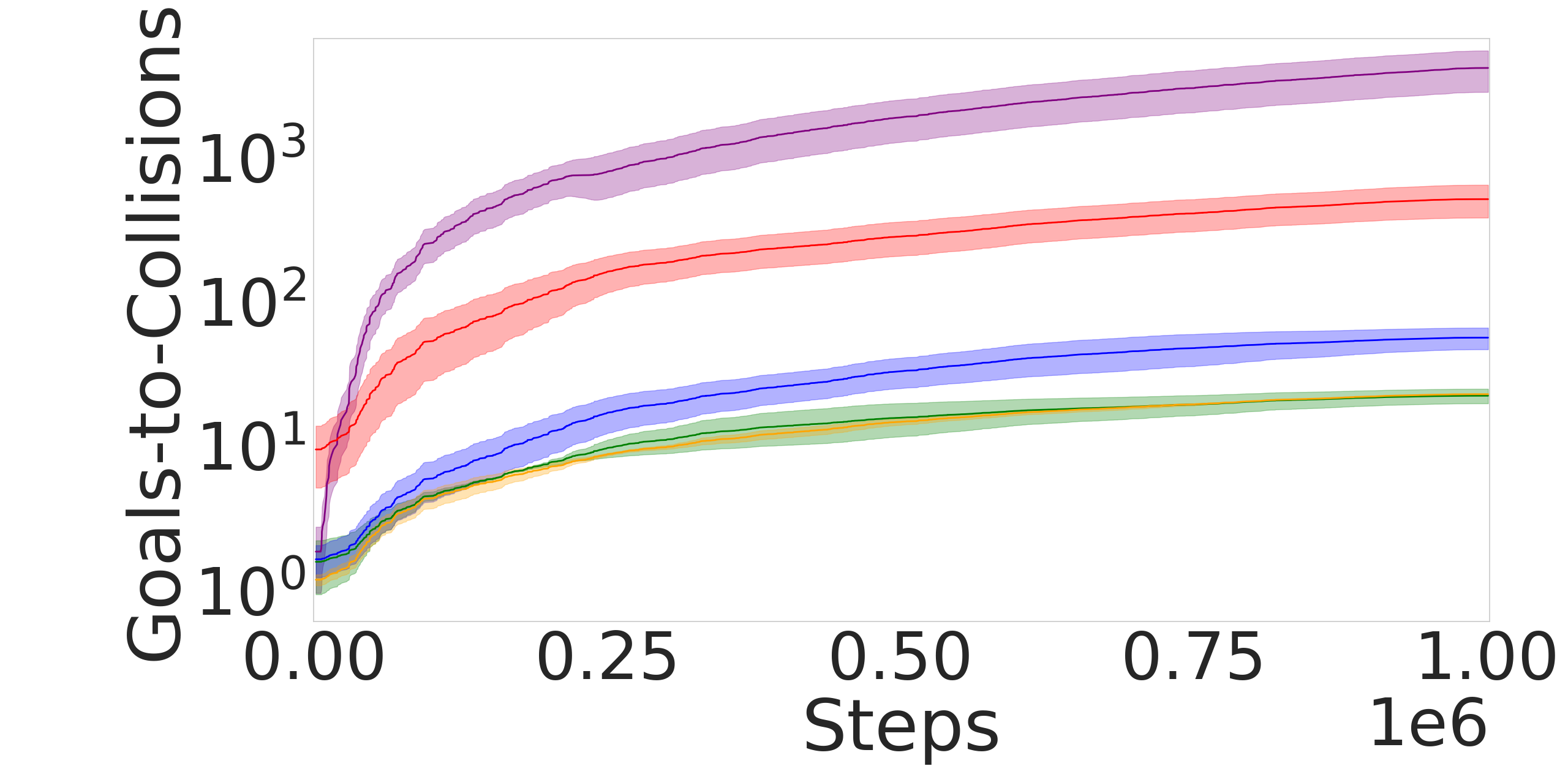}
        \label{fig:ratio_1}
    }
    \hfill
    \subfigure[]{
        \includegraphics[width=0.315\textwidth, trim={35pt 0pt 55pt 0pt}, clip]{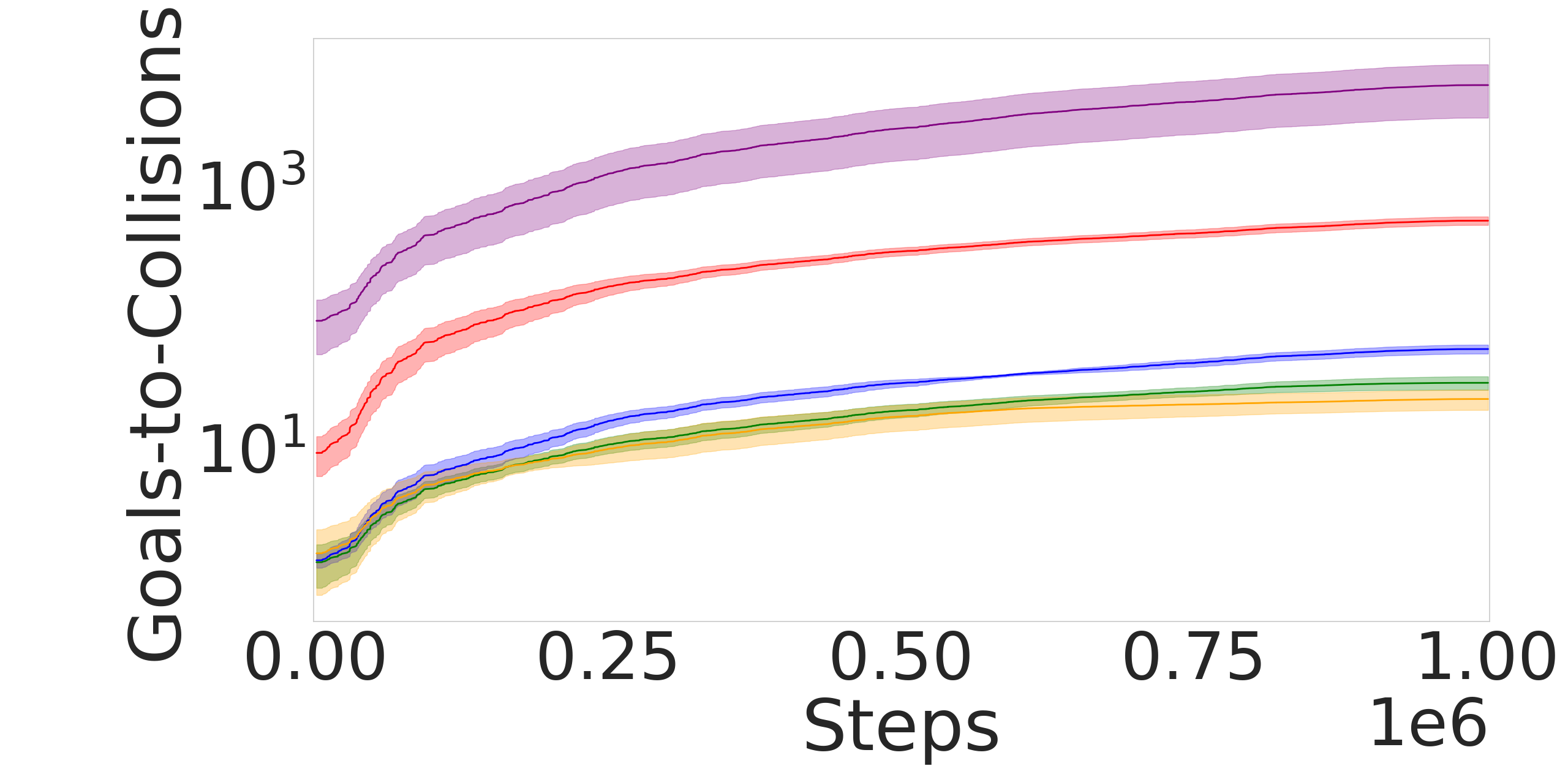}
        \label{fig:ratio_2}
    }
    \vspace{3pt} 
    \textbf{Accumulated Goals to Accumulated Collisions Ratios}
    \vspace{0.5em} 

    \subfigure[]{
        \includegraphics[width=0.315\textwidth, trim={35pt 0pt 55pt 0pt}, clip]{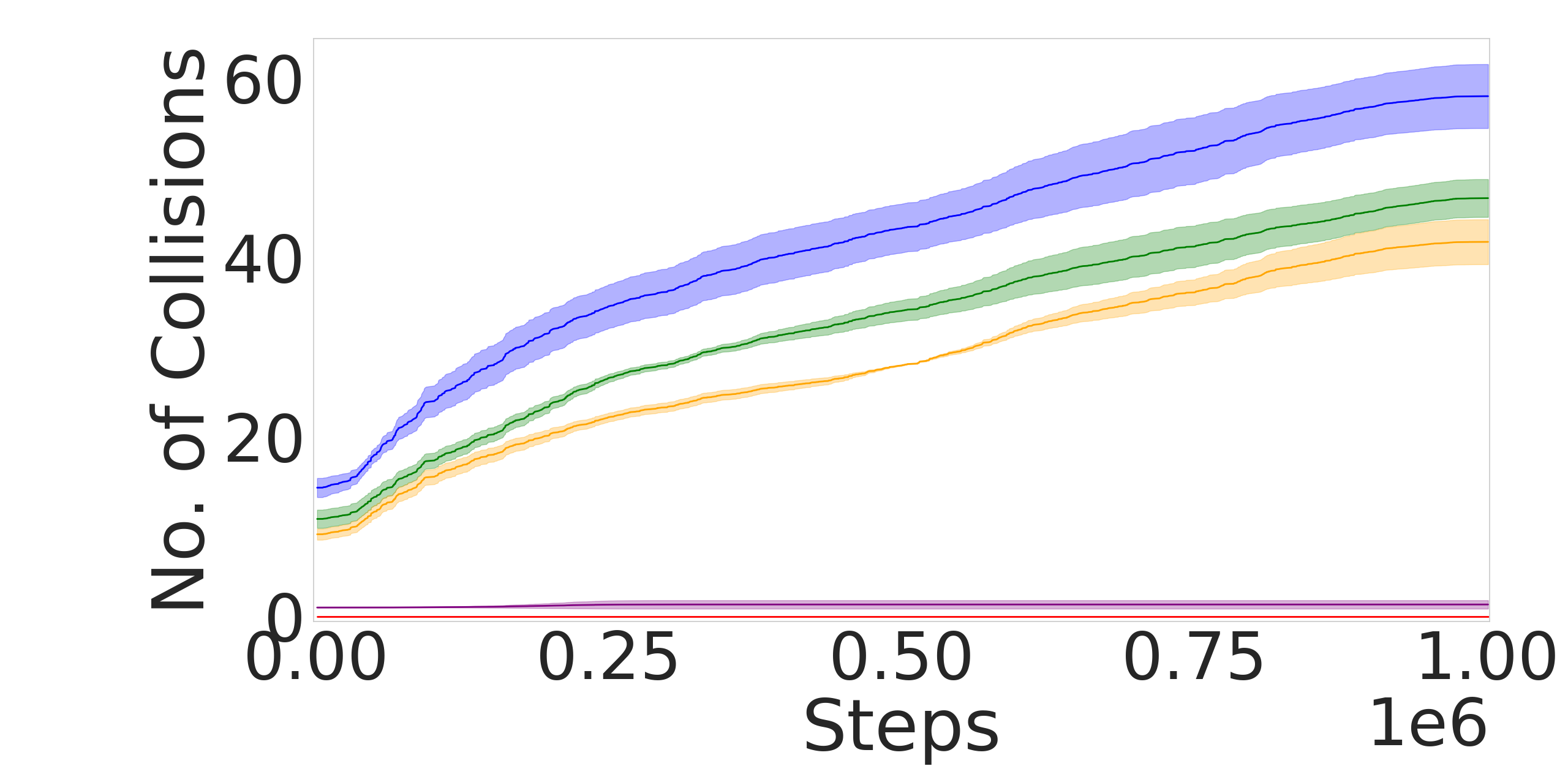}
        \label{fig:cols_0}
    }
    \hfill
    \subfigure[]{
        \includegraphics[width=0.315\textwidth, trim={35pt 0pt 55pt 0pt}, clip]{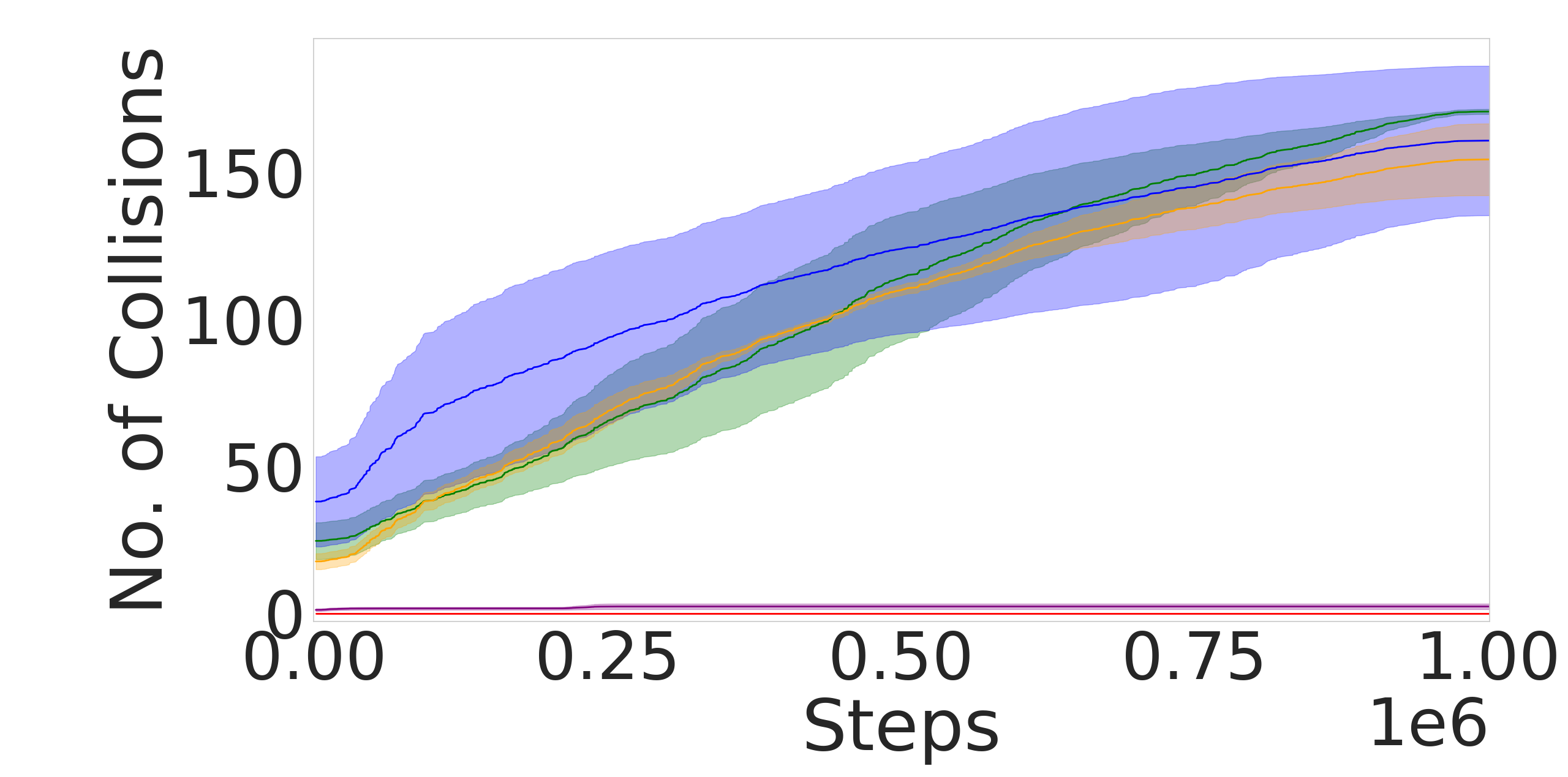}
        \label{fig:cols_1}
    }
    \hfill
    \subfigure[]{
        \includegraphics[width=0.315\textwidth, trim={35pt 0pt 55pt 0pt}, clip]{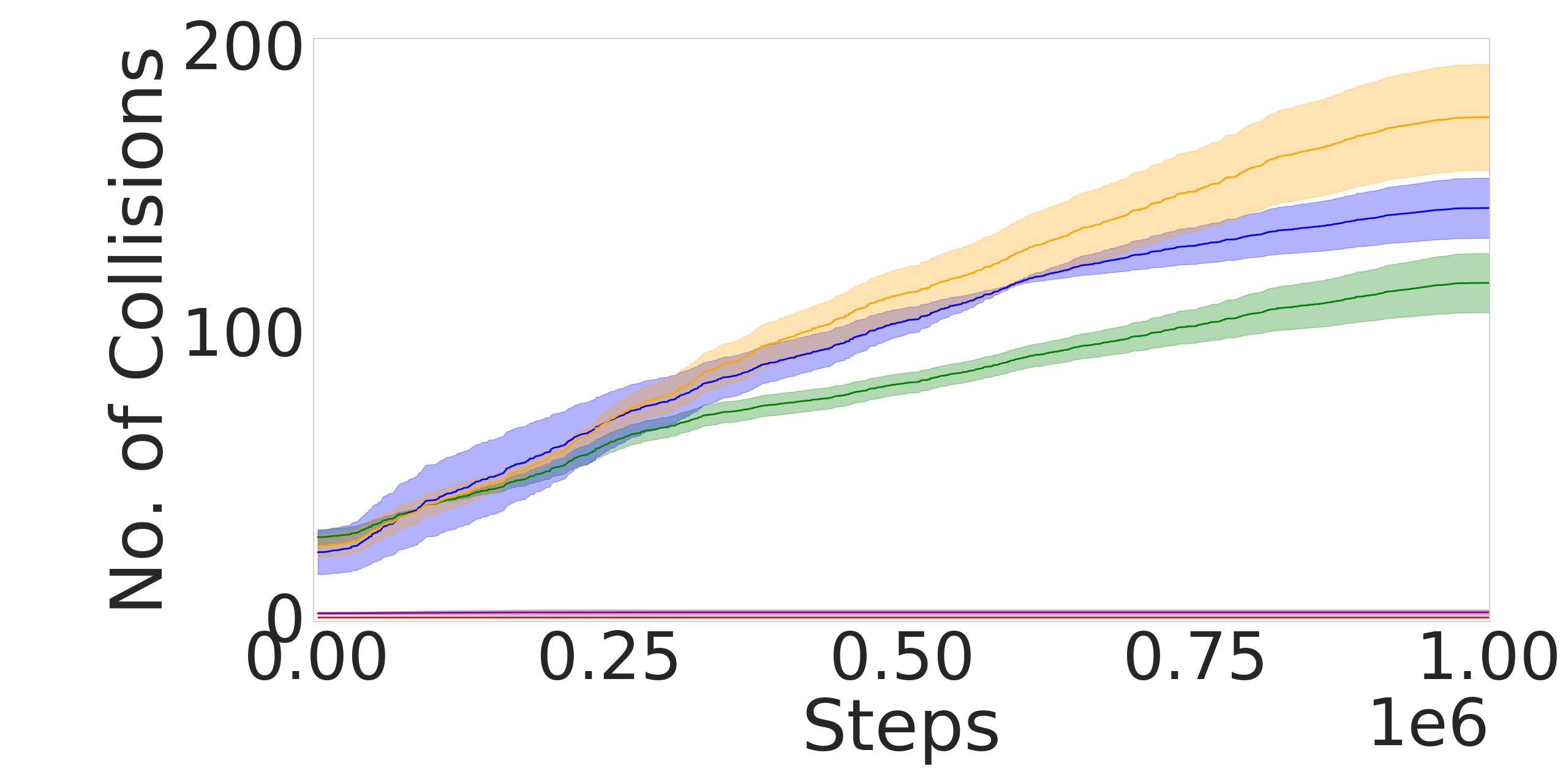}
        \label{fig:cols_2}
    }
    \vspace{3pt} 
    \textbf{Total Number of Collisions}
    \vspace{0.5em} 

    \subfigure[]{
        \includegraphics[width=0.315\textwidth, trim={35pt 0pt 55pt 0pt}, clip]{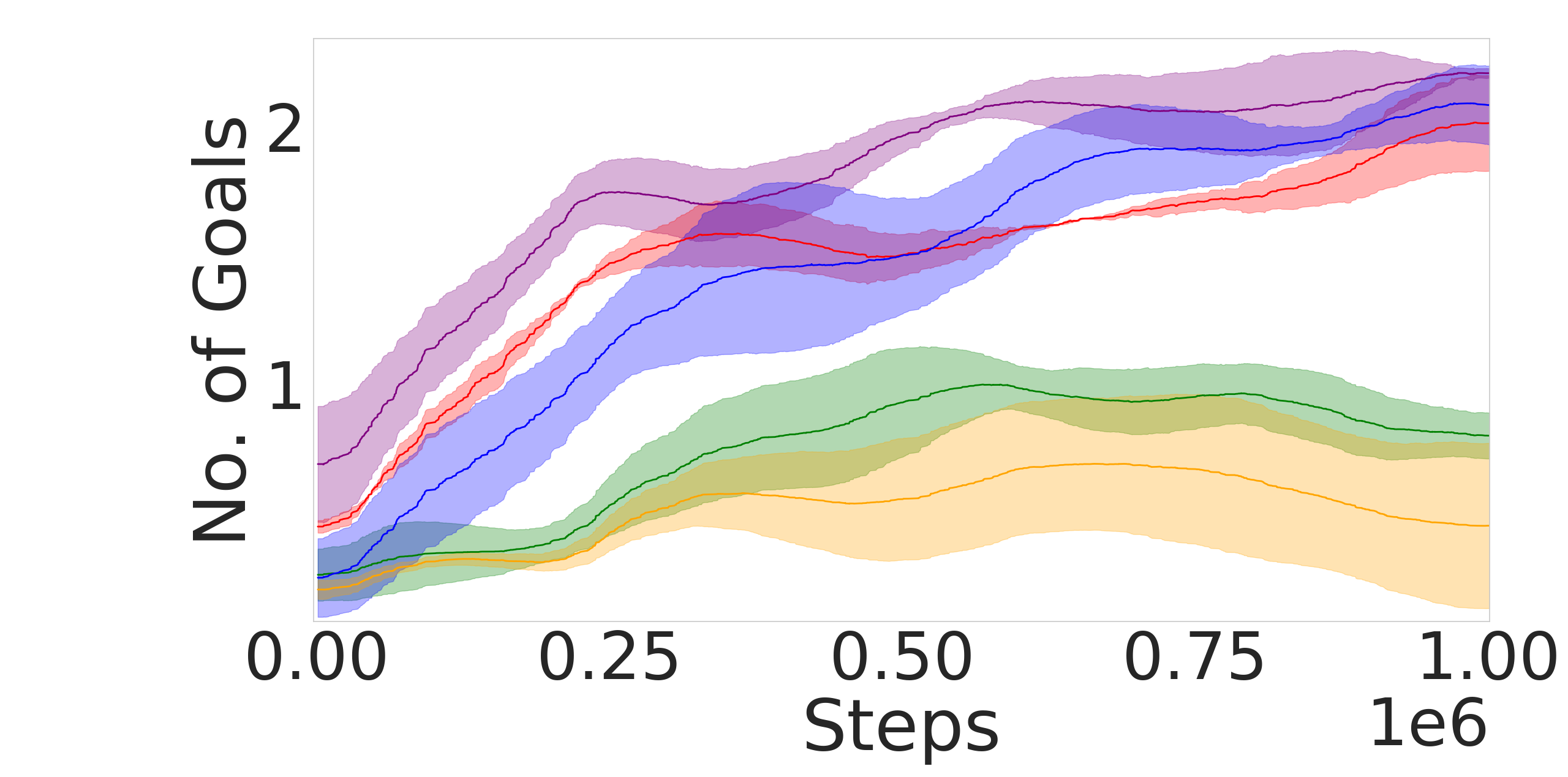}
        \label{fig:goals_0}
    }
    \hfill
    \subfigure[]{
        \includegraphics[width=0.315\textwidth, trim={35pt 0pt 55pt 0pt}, clip]{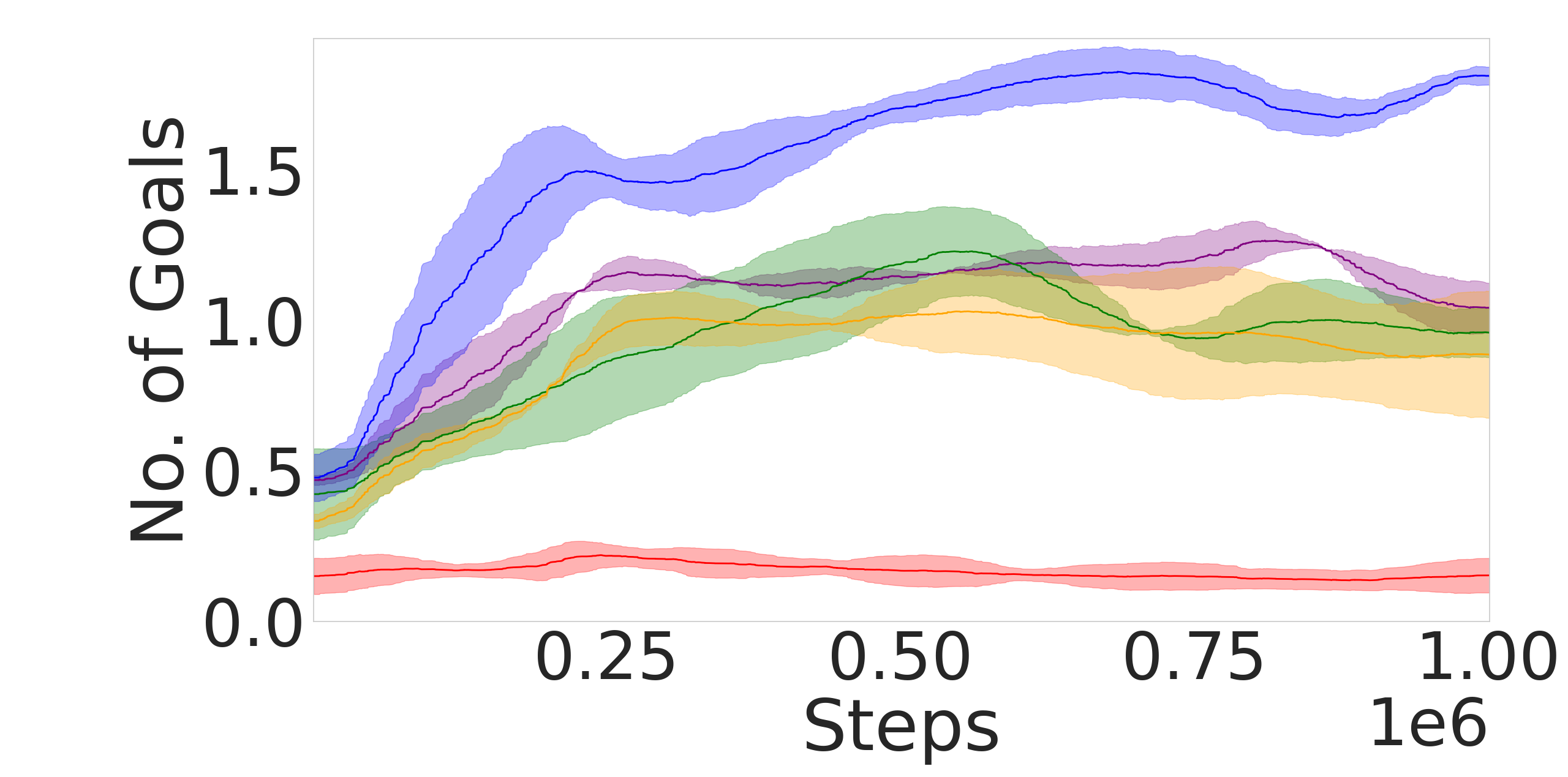}
        \label{fig:goals_1}
    }
    \hfill
    \subfigure[]{
        \includegraphics[width=0.315\textwidth, trim={35pt 0pt 55pt 0pt}, clip]{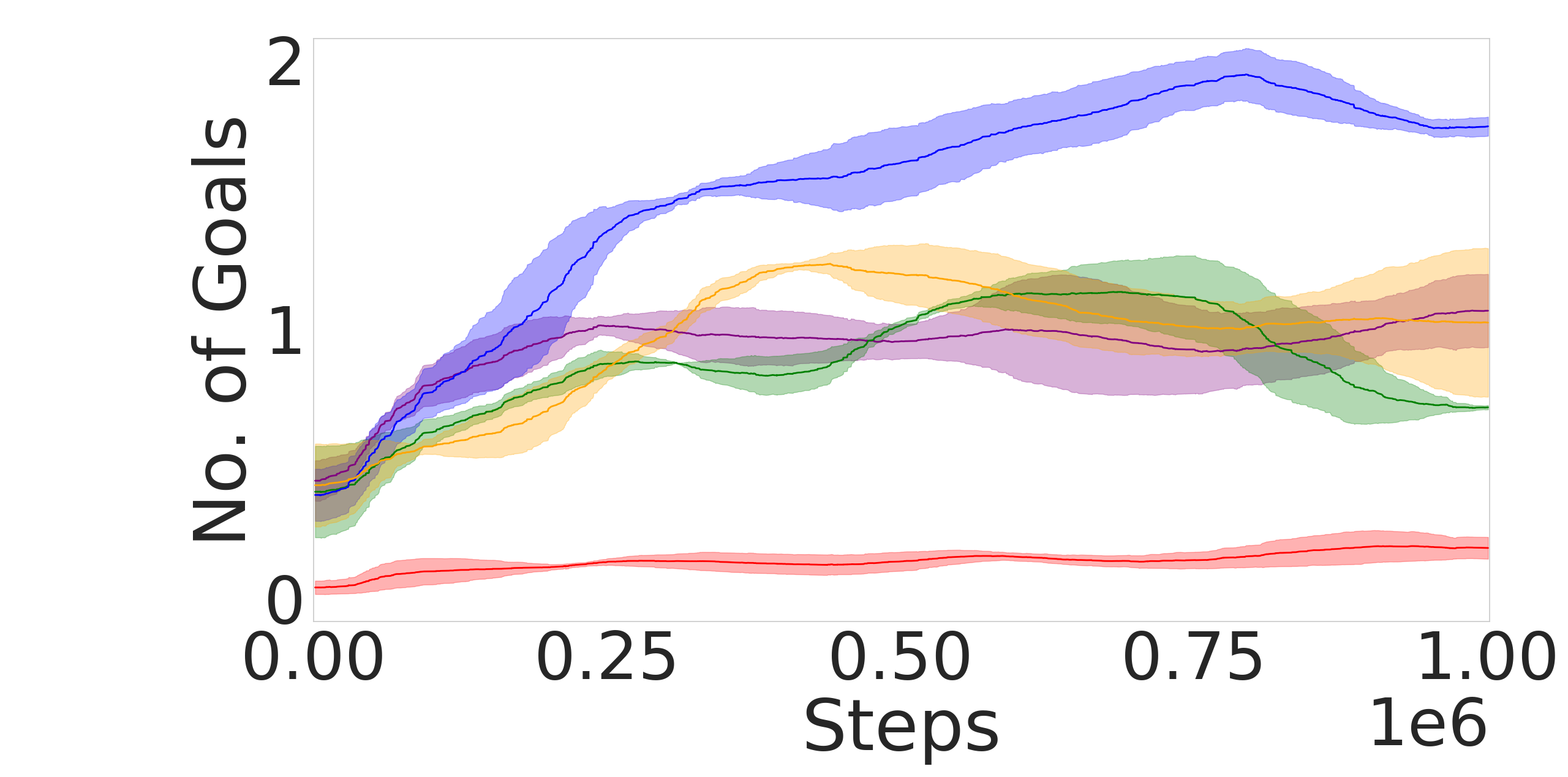}
        \label{fig:goals_2}
    }
    \textbf{Number of Reached Goals per Training Step}
    \caption{\small
        Results for all the approaches in the three environments. The bold lines show the average of three random seeds, while the shaded areas show the standard deviation over the runs. Our approach consistently achieves the highest goals-to-collisions ratio, which shows that our approach is able to safely guide the task agent to learn the navigation task with a few number of collisions. Note that some baselines reach up to 200 collisions, while our approach achieves near-zero collisions, overlapping with the MPC\_TUNED at zero, as can be seen in the second row.
    }
    \label{fig:results}
    \vspace{-20pt}
\end{figure}
\vspace{-8pt}
\subsection{Experimental Setup}
We carry out the experiments in PyBullet [\cite{coumans2021}] in three different environments. The first environment, Fig.\ref{fig:env0}, contains five cylindrical obstacles (pillars) placed randomly in the environment at the beginning of each episode, the second environment consists of six pillars and six L-shaped walls, Fig.\ref{fig:env1}, while the third consists of eight pillars and eight L-shaped walls, Fig.\ref{fig:env2}. We chose the pillars as they are commonly used in obstacle avoidance scenarios, and the L-shaped walls as they represent challenging local minima during trajectory planning. 
During training, the RL agents, the obstacles locations, goals, and the robot location are randomized at the beginning of each episode. We apply episodic training, where the episode terminates if the robot collides or if it is not moving for 30 consecutive steps, or the maximum number of steps is reached. The robot is capable of reaching multiple goals within a single episode; a new goal is randomly generated each time the previous one is reached.

To train the supervisor agent while minimizing the number of collisions, we found it beneficial to repeatedly add samples involving collisions to its corresponding replay buffer. Empirically, duplicating each such sample three times provided the best results. Additionally, we train the supervisor agent more frequently compared to the task agent. Having different agents with separate replay buffers, makes it possible to train the agents for different number of steps without destabilizing the training. Finally, we found that annealing the learning rates for the supervisor agent is also extremely beneficial for achieving less number of collisions.

All the agents have been trained for one million steps from scratch. To assess the performance of the agents, we use three metrics as follows: \textbf{(i) Accumulated Number of Goals-to-Accumulated Number of Collisions Ratio:} This metric has been introduced in [\cite{thananjeyan2021recovery}] to assess the performance of safe RL approaches as it signifies the importance of achieving more goals and less collisions. A higher ratio indicates fewer collisions while achieving more goals, which is desirable. In case of zero collisions, we set the number of collisions to one, to avoid division by zero. \textbf{(ii) Accumulated Number of Collisions:} Additionally, we count the total number of collisions during the training, as this is the main focus of safe RL. Fewer collisions indicate better safety adherence. \textbf{(iii) Number of Reached Goals:} Finally, we keep track of the number of goals reached at each training step to indicate the progress of the task agent. A higher number indicates better behavior from the task agent.
\begin{figure}[t]
    \centering
        \textcolor{matplotlibpurple}{\rule[0.5ex]{0.5cm}{4pt}} Ours \hspace{1em}
        \textcolor{matplotliborange}{\rule[0.5ex]{0.5cm}{4pt}} Ours+Goal Information \hspace{1em}

    \subfigure[Goals to Collisions Ratio]{
        \includegraphics[width=0.315\textwidth, trim={33pt 0pt 55pt 0pt}, clip]{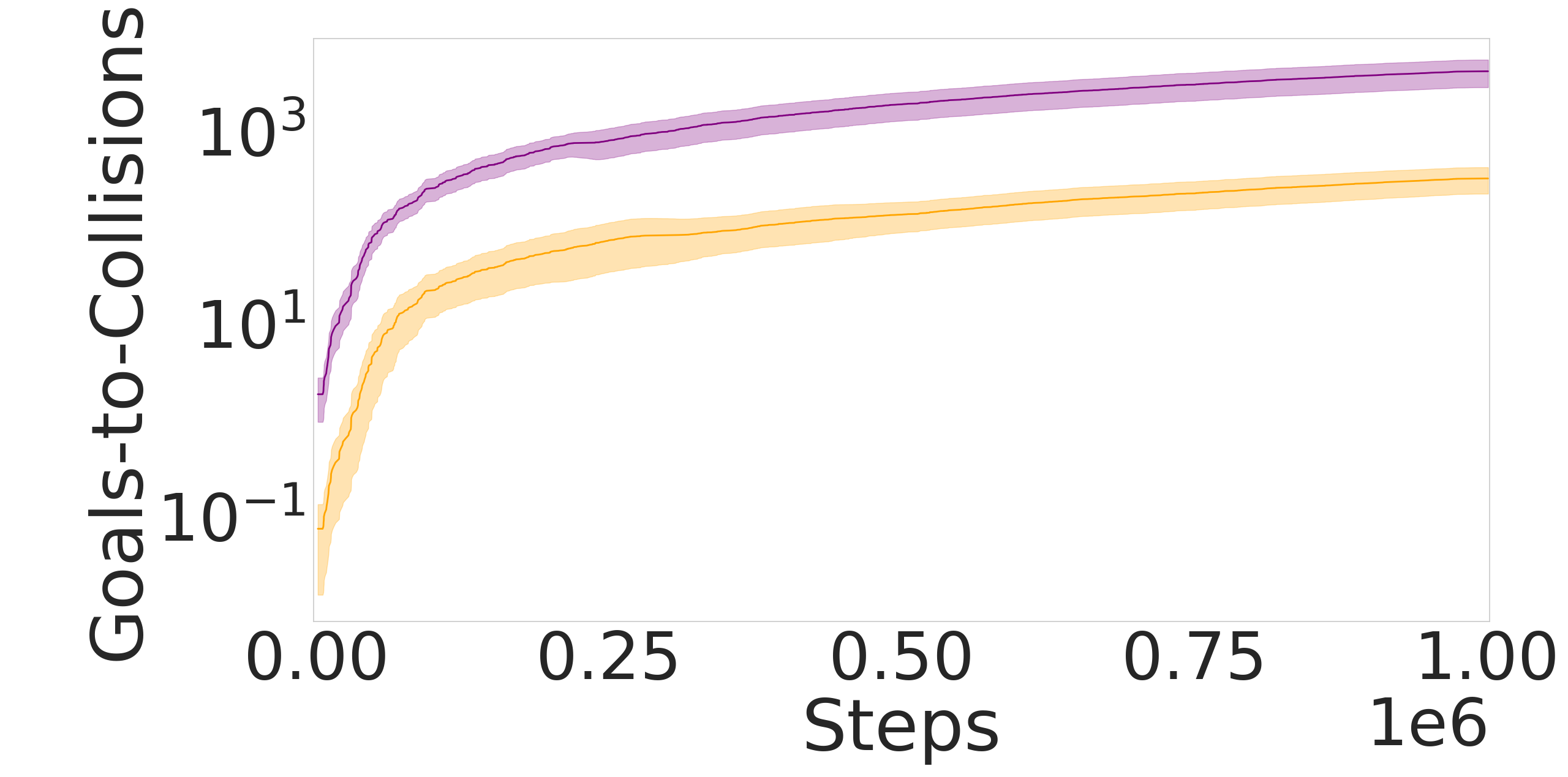}
        \label{fig:abl_ratio}
    }
    \hfill
    \subfigure[Total Number of Collisions]{
        \includegraphics[width=0.315\textwidth, trim={35pt 0pt 55pt 0pt}, clip]{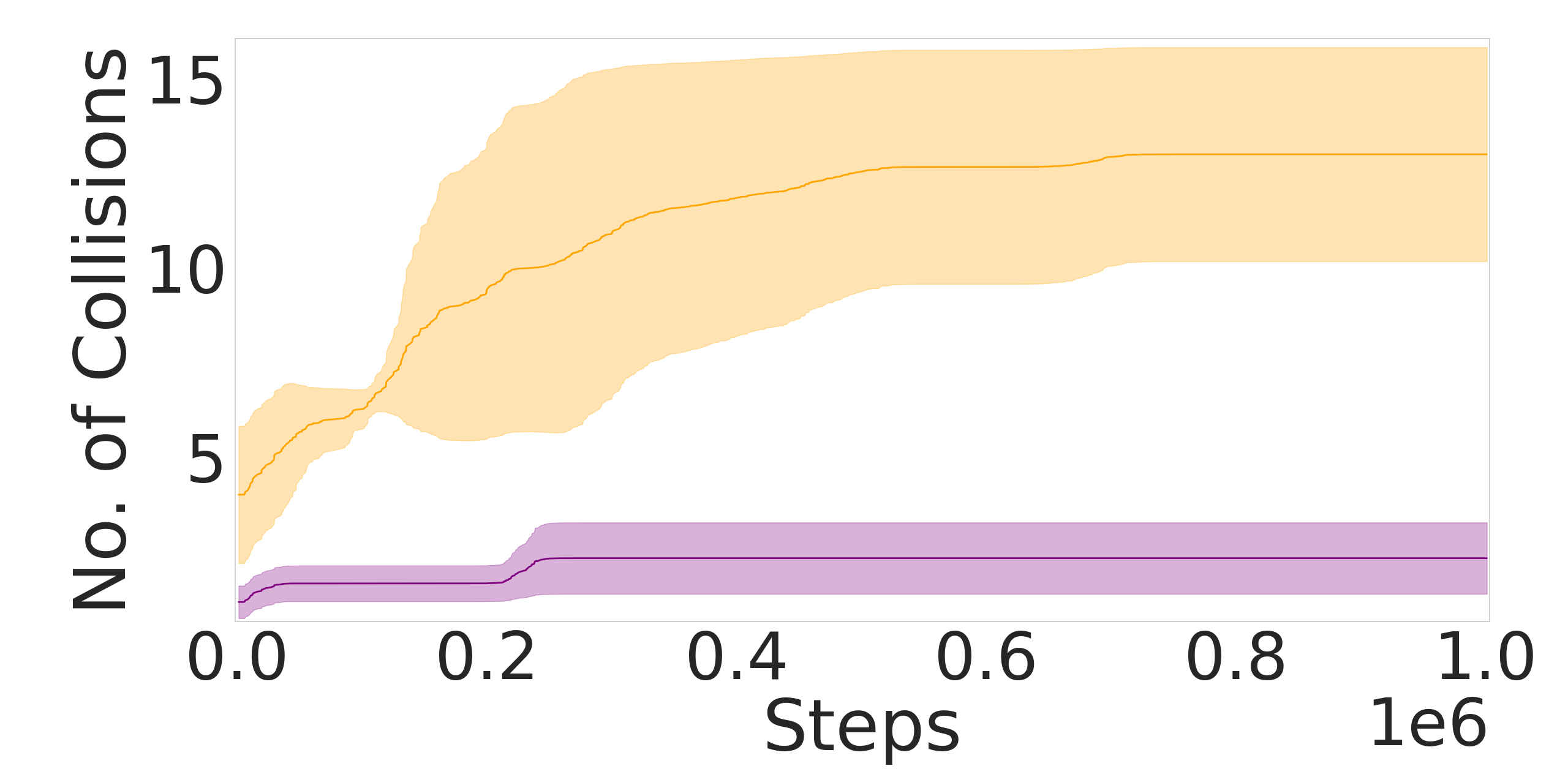}
        \label{fig:abl_cols}
    }
    \hfill
    \subfigure[Number of Reached Goals]{
        \includegraphics[width=0.315\textwidth, trim={35pt 0pt 55pt 0pt}, clip]{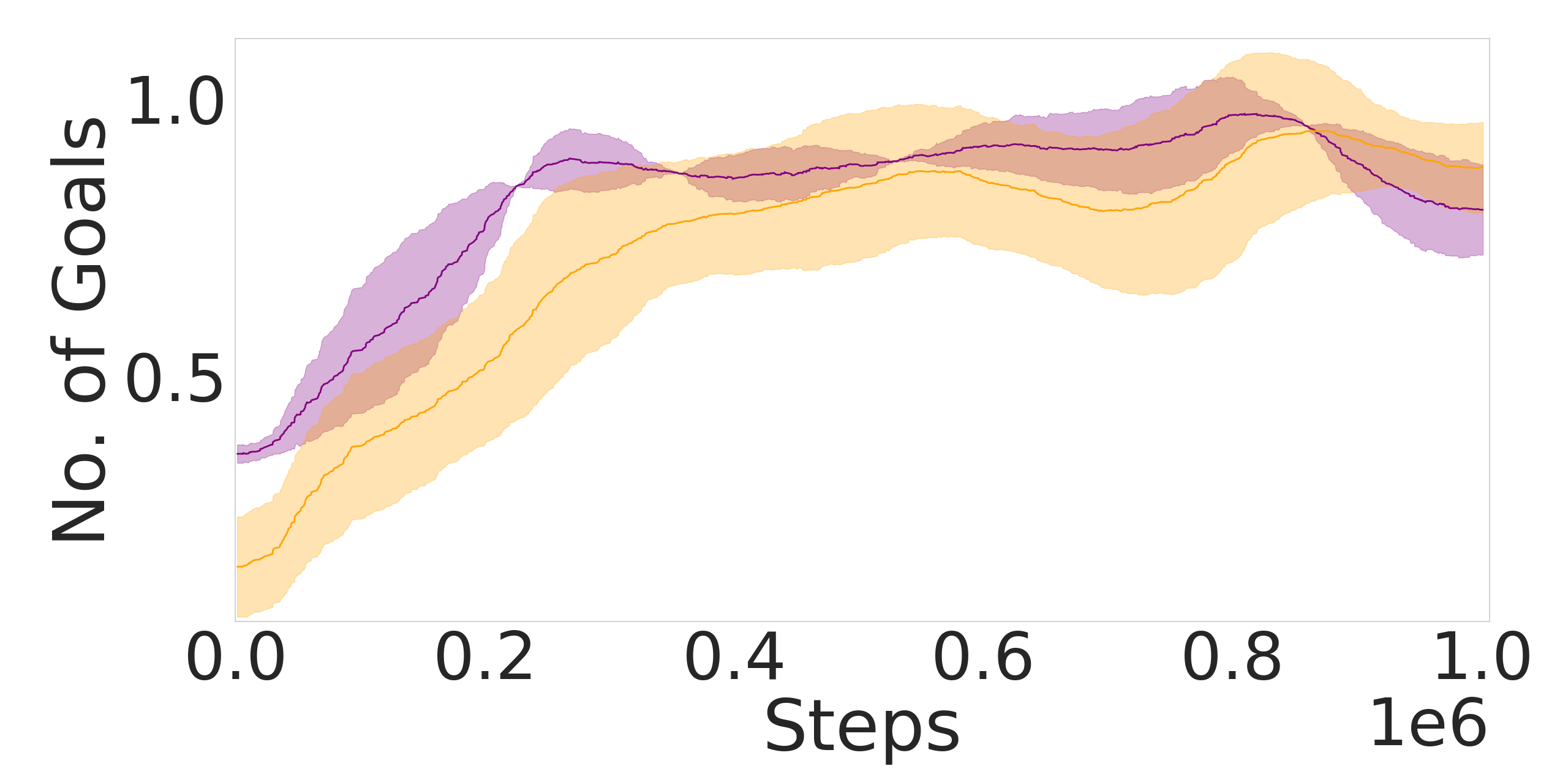}
        \label{fig:abl_goals}
    }
    \caption{\small
        Results for the ablation study over three random seeds. Introducing the goal information to the supervisor agent, results in more collisions as the supervisor agent explores to reach more goals to maximize its rewards. The goals-to-collisions ratio for our approach without the goal information is higher than for the agent with the goal information. This shows that including the goal information for the supervisor agent does not improve the performance of the task agent.
    }
    \label{fig:abl_res}
        \vspace{-20pt} 
\end{figure}
\vspace{-10pt}
\subsection{Results}
Figure \ref{fig:results} shows the results of the three metrics for all the methods in the three environments. The figures show the means (bold lines) and standard deviations (shaded areas) over three runs for each approach. Figures \ref{fig:ratio_0}, \ref{fig:ratio_1}, and \ref{fig:ratio_2} illustrate that our approach achieves the highest goals-to-collisions ratio in all the environments compared to the baselines. In the first environment, MPC\_TUNED has a high ratio as the MPC shield is able to eliminate the collisions completely while navigation to the goals. In the more challenging environments, the MPC\_TUNED still maintains zero collisions, see Fig. \ref{fig:cols_0}, \ref{fig:cols_1}, and \ref{fig:cols_2}  but at the cost of over constraining the task agent resulting in the least number of goals reached among all the baselines. The unconstrained SAC is able to reach the highest number of goals, Fig. \ref{fig:goals_0}, \ref{fig:goals_1}, and \ref{fig:goals_2} in all environments at the cost of large number of collisions, resulting in low accumulated goals-to accumulated collisions ratios. The constrained RL approaches SAC\_LAG and SAC\_PID on the other hand are able to decrease the number of collisions compared to the SAC at the cost of less goals reached, consequently leading to low ratios as well. These results demonstrate our method's ability to minimize collisions without over constraining the task agent.
\vspace{-5pt}
\subsection{Ablation Study}

As mentioned previously, we do not provide any task information to the supervisor agent to minimize the exploration of the supervisor agent and hence minimize the number of collisions. In this section, we perform an ablation study in the first environment, where we add the goal information to the supervisor agent and give it a reward for reaching the goal and compare our approach without the goal information against the agent with the goal information. As can be seen from Fig. \ref{fig:abl_res}, introducing the goal information increases the number of collisions, as the added information expands the state space for the supervisor agent, resulting in a slower training process. This, in turn, reduces the goals-to-collisions ratio compared to our approach without the additional information. Nevertheless, the number of collisions remained lower than the constrained RL baselines. 
\vspace{-5pt}
\subsection{Real-World Experiment}

\begin{figure}[t!]
    \centering
    \begin{minipage}{0.46\textwidth} 
        \centering
        \includegraphics[width=\textwidth]{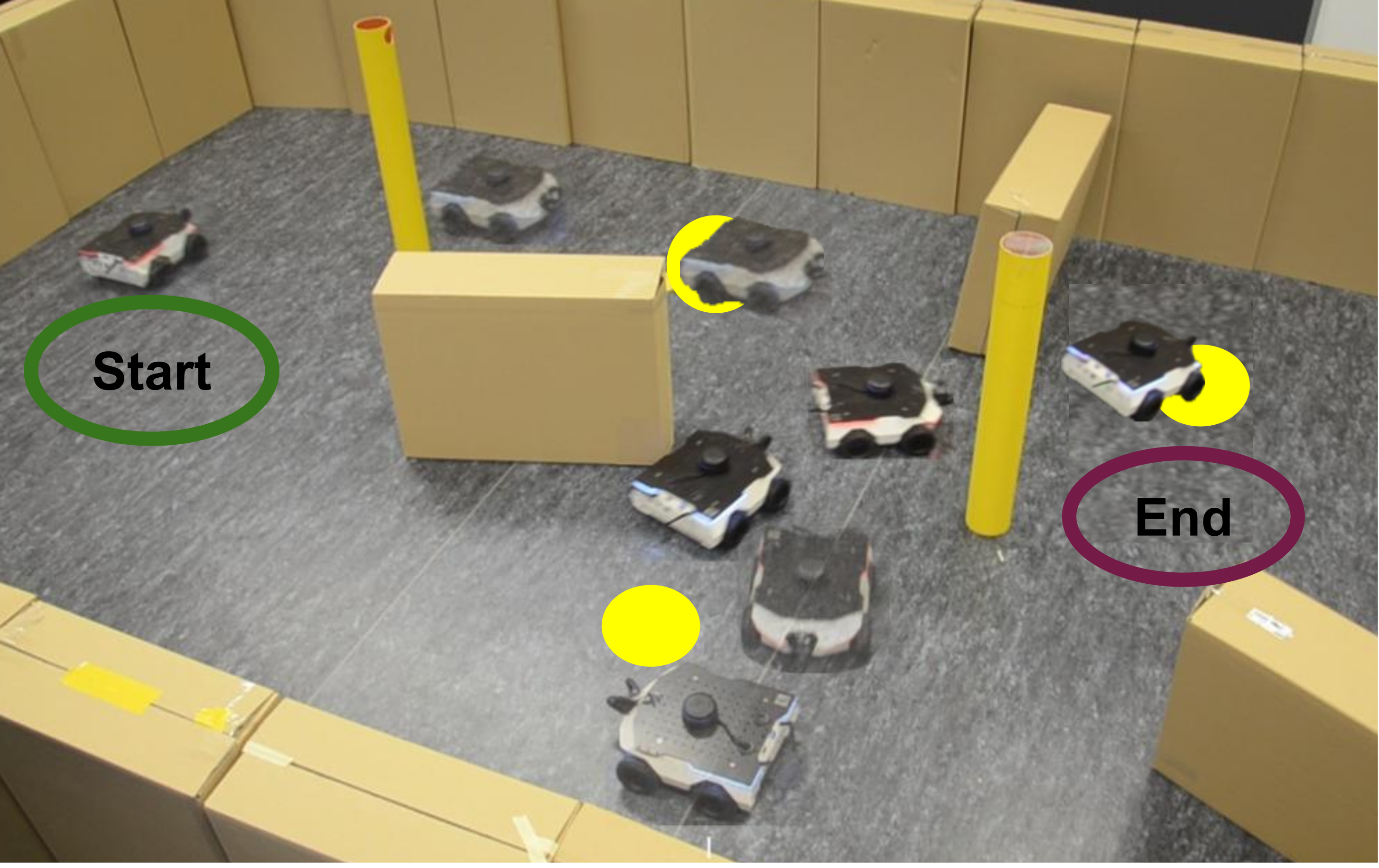}
        \caption*{\centering Robot Trajectory}
        \label{fig:large_fig}
    \end{minipage}%
    \hfill
    \begin{minipage}{0.53\textwidth} 
        \centering
        \subfigure[Weight 1]{%
            \includegraphics[width=0.48\textwidth]{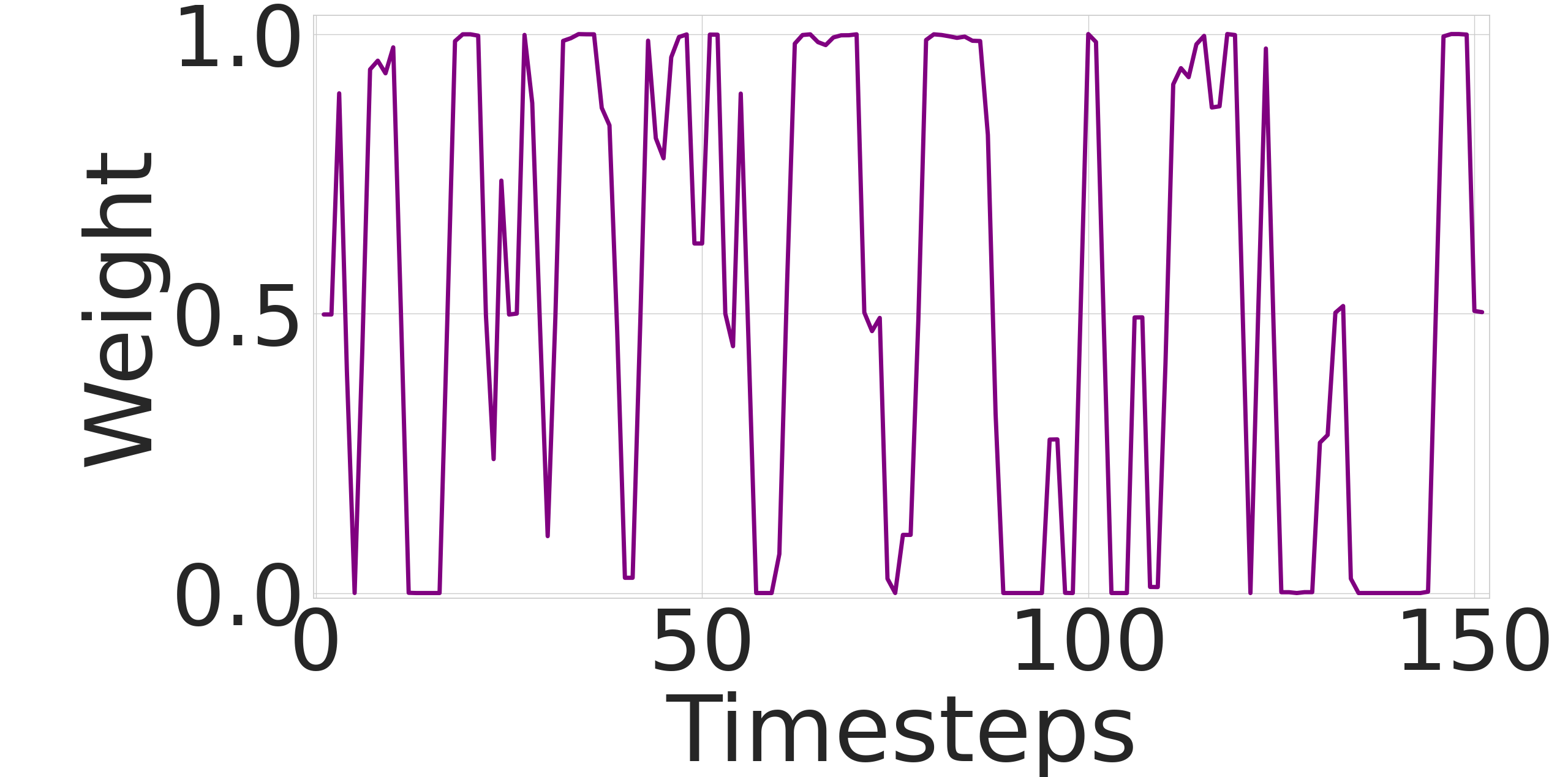}
            \label{fig:wt1}
        }
        \subfigure[Weight 4]{%
            \includegraphics[width=0.48\textwidth]{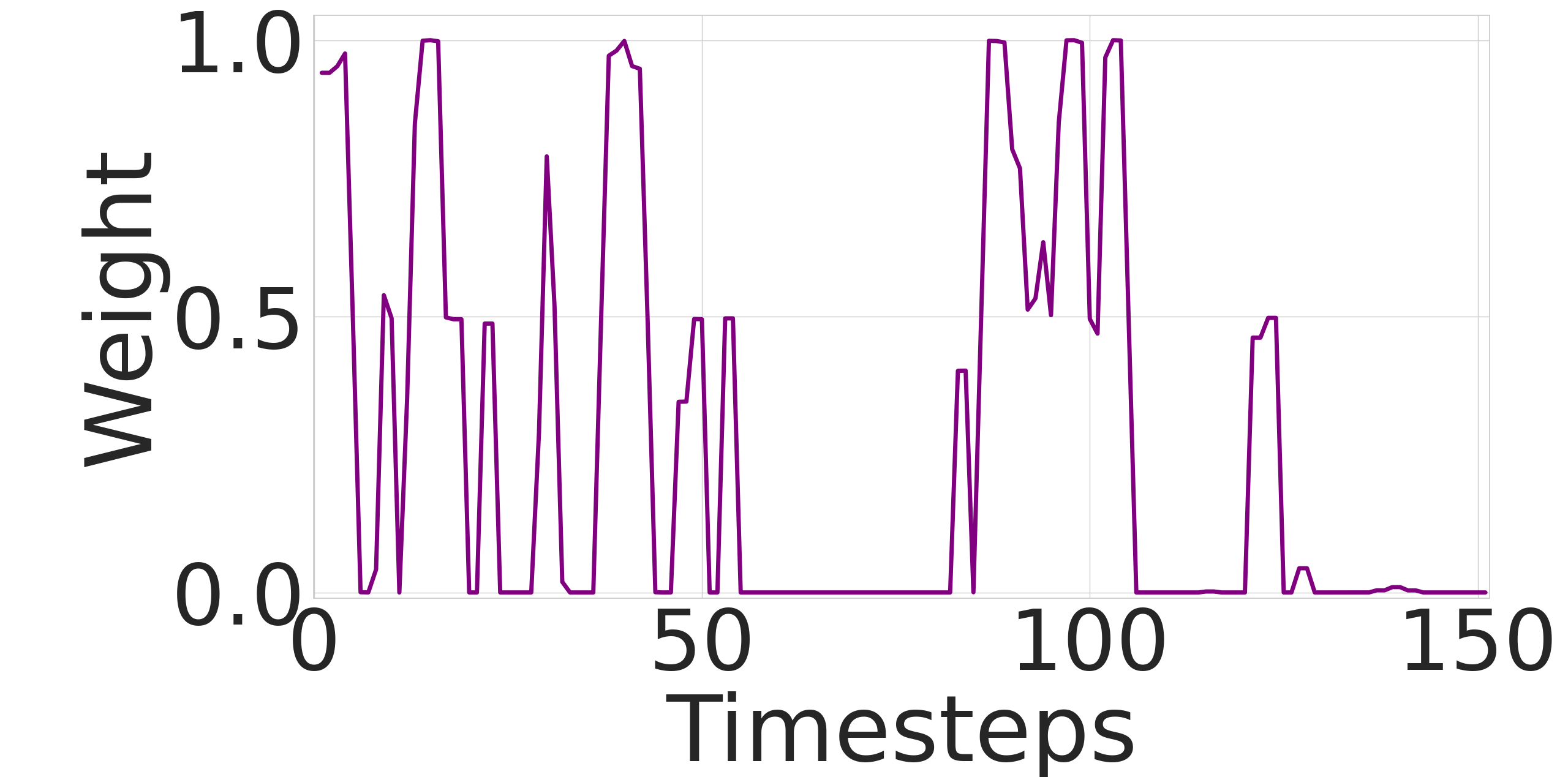}
            \label{fig:wt4}
        } \\[5pt]
        \subfigure[Weight 2]{%
            \includegraphics[width=0.48\textwidth]{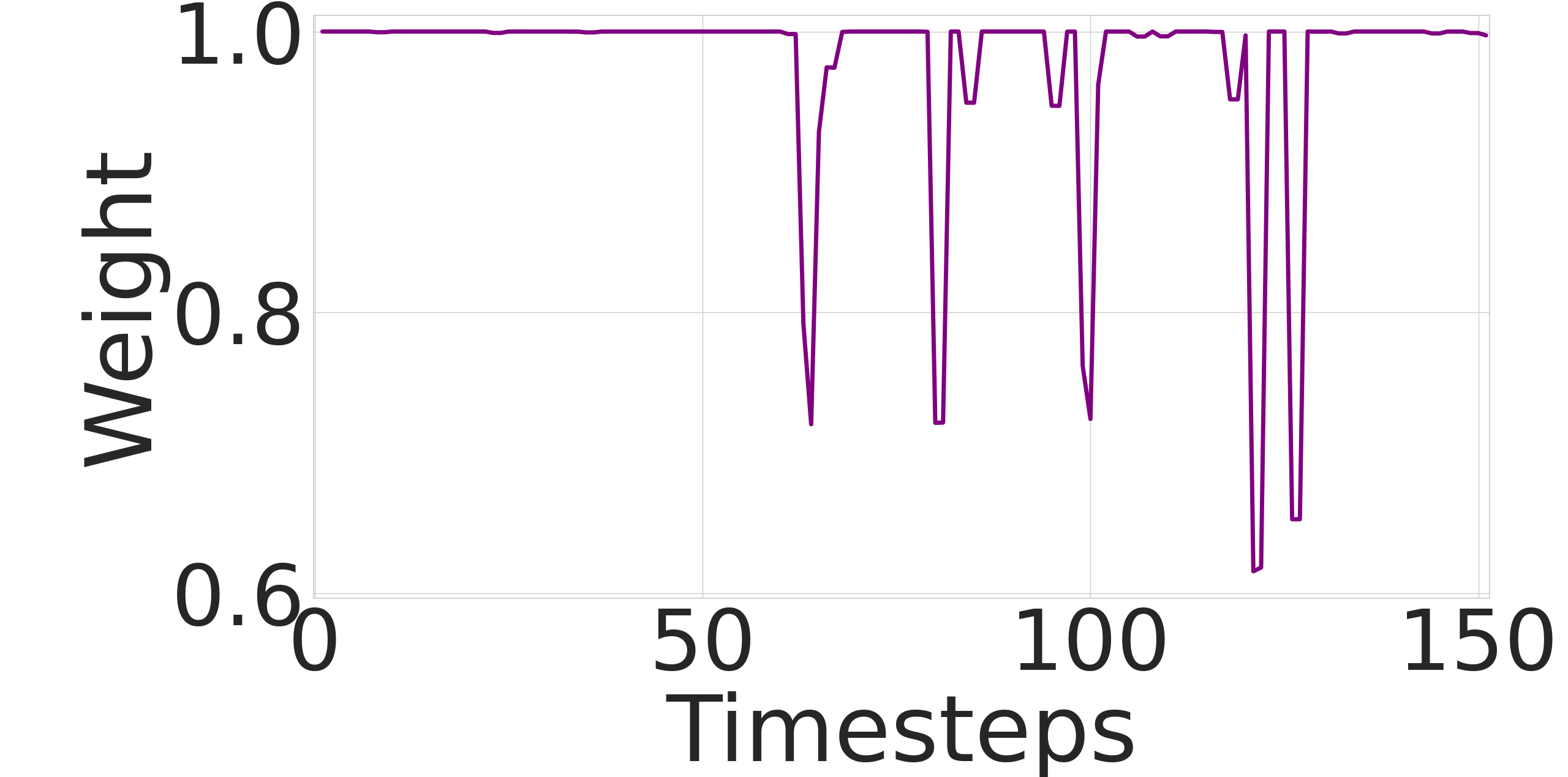}
            \label{fig:wt2}
        }
        \subfigure[Weight 3]{%
            \includegraphics[width=0.48\textwidth]{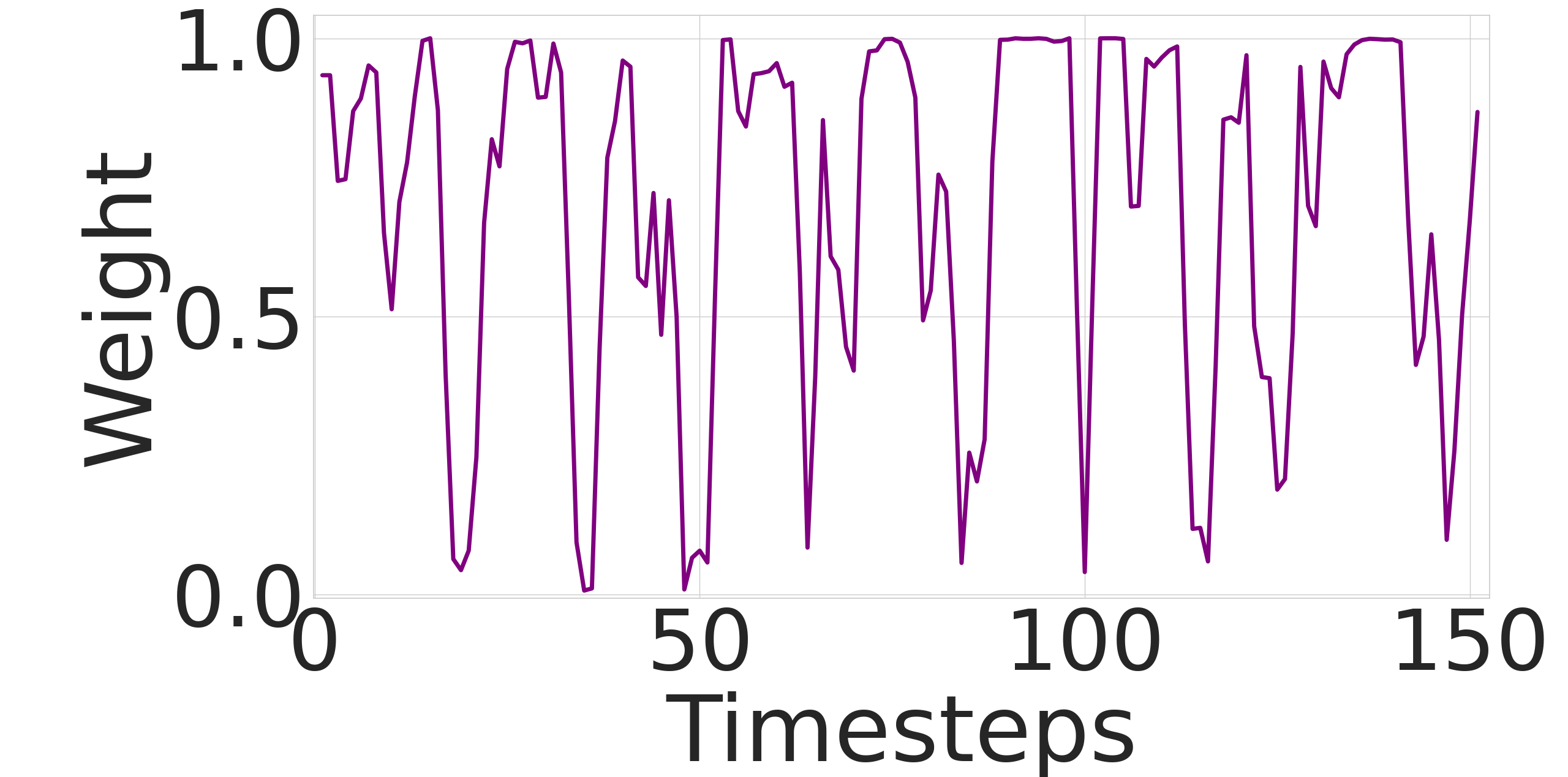}
            \label{fig:wt3}
        }
    \end{minipage}

    \caption{\small The figure illustrates the real-robot trajectory (left) and the weights adjusted by the supervisor agent~(right). The robot navigates from the start position to the goals (yellow circles) while avoiding obstacles. The weight plots are arranged such that each weight corresponds to its respective quadrant in the lidar data. It can be seen that the supervisor agent increases the weights of the obstacles as the robot moves closer to them, prioritizing obstacle avoidance in the MPC shield. }
    \label{fig:real_exp}
\vspace{-20pt}
\end{figure}

Finally, we investigate how the supervisor agent modifies the weights for the obstacles in a real-world experiment. Using checkpoints from our trained policies for both the task agent and the supervisor agent, we deployed these policies on a real robot. The policies were originally trained in simulation at 5 Hz, so we maintained the same control frequency on the robot to ensure consistency. We used the ROSbot XL \footnote{https://husarion.com/manuals/rosbot-xl/overview/ } from Husarion for the experiments. The robot is equipped with a $360^\circ$ lidar and an active tracker to track its location. See Fig.~\ref{fig:real_exp}.

The supervisor agent, implemented as an RL policy, dynamically adjusts the weights of the obstacles based on the current environment and state information. These weights influence the importance assigned to specific obstacles when calculating safe actions. For instance, as the robot moves closer to an obstacle, the supervisor agent increases the corresponding weight, prioritizing obstacle avoidance in the MPC shield. Conversely, weights for farther or less relevant obstacles may decrease, allowing the robot to focus on reaching its goal efficiently.

Fig.\ref{fig:real_exp} illustrates the robot’s trajectory as it navigates toward marked goals while avoiding obstacles. The plot shows the real-time changes in weights as modified by the supervisor agent. These changes reflect the RL agent’s decision-making process to balance obstacle avoidance with the task agent’s goal-directed actions.
\vspace{-10pt}

\section{Conclusion}
This work presents a novel safety shield approach for reinforcement learning (RL) in navigation tasks, designed to effectively balance safety and exploration. By combining the robustness of model predictive control (MPC) safety shields with the long predictive capabilities of RL, we introduced a dynamic system where a supervisor agent adjusts the weights for the obstacle avoidance terms and for aligning the MPC actions with the task agent's actions.

Our experiments demonstrate the superiority of the proposed approach across diverse environments. The results show that our method achieves the highest goals-to-collisions ratio, significantly minimizing collisions compared to constrained RL methods, i.e., \cite{stooke2020responsive, ray2019benchmarking}. Unlike classical MPC-based shields \cite{dawood2025ral}, which often over-constrains the RL agents, our method promotes exploration without compromising safety, enabling the task agent to achieve higher rewards in challenging scenarios. Moreover, compared to unconstrained RL methods, which maximize goals at the cost of collisions, our approach strikes a balance by reducing collisions while maintaining competitive performance in goal-reaching metrics.

The effectiveness of the dynamic adjustment mechanism was evident in its ability to adapt to varying environmental complexities, ensuring fewer constraints violations without hindering the task agent's progress. This balance emphasizes the potential of integrating optimization-based methods with RL to address real-world challenges in safe exploration.

\section*{Acknowledgments}
Murad Dawood, Ahmed Shokry, and Maren Bennwitz are with the Humanoid
Robots Lab, University of Bonn, Germany and with the Lamarr Institute
for Machine Learning and Artificial Intelligence as well as the Center for
Robotics, Bonn, Germany. This work has been partially funded by the
BMBF within the Robotics Institute Germany, grant No. 16ME0999.

\bibliography{bibliography}

\end{document}